\documentclass[10pt,twocolumn,letterpaper]{article}

\usepackage{cvpr}              
\usepackage{soul}
\usepackage{float}
\usepackage{booktabs}
\usepackage[table]{xcolor}

\definecolor{latblue}{HTML}{D9E8FF}     
\definecolor{tradeyellow}{HTML}{FFF2B2} 
\definecolor{errgreen}{HTML}{DFF2D8}    

\definecolor{cvprblue}{rgb}{0.21,0.49,0.74}
\usepackage[pagebackref,breaklinks,colorlinks,allcolors=cvprblue]{hyperref}
\usepackage[table]{xcolor}   
\usepackage{caption}
\usepackage{placeins}
\usepackage{colortbl}   
\usepackage[accsupp]{axessibility}  


\usepackage{subcaption}
\usepackage{graphicx}
\def\paperID{3} 
\def\confName{CVPR}
\def\confYear{2026}

\title{EdgeVTP: Exploration of Latency-efficient Trajectory Prediction for Edge-based Embedded Vision Applications}

\author{%
  \begin{tabular}[t]{@{}c@{\hspace{2em}}c@{}}
    \begin{tabular}[t]{c}
      Seungjin Kim\\
      University of Wyoming\\
      {\tt\small skim27@uwyo.edu}
    \end{tabular} &
    \begin{tabular}[t]{c}
      Reza Jafarpourmarzouni\\
      University of North Carolina at Charlotte\\
      {\tt\small rjafarpo@charlotte.edu}
    \end{tabular}\\[1.35ex]
    \begin{tabular}[t]{c}
      Christopher Neff\\
      North Carolina A\&T State University\\
      {\tt\small cgneff@ncat.edu}
    \end{tabular} &
    \begin{tabular}[t]{c}
      Hamed Tabkhi\\
      University of North Carolina at Charlotte\\
      {\tt\small htabkhiv@charlotte.edu}
    \end{tabular}\\[1.35ex]
    \multicolumn{2}{c}{%
      \begin{tabular}[t]{c}
        Vinit Amrutlal Katariya\\
        University of Wyoming\\
        {\tt\small vkatariy@uwyo.edu}
      \end{tabular}%
    }
  \end{tabular}%
}

\begin{document}
\maketitle
\begin{abstract}
Vehicle trajectory prediction is central to highway perception, but deployment on roadside edge devices necessitates bounded, deterministic end-to-end latency. We present EdgeVTP, an embedded-first trajectory predictor that combines interaction-aware graph modeling with a lightweight transformer backbone and a one-shot curve decoder. By predicting future motion as compact curve parameters (anchored at the last observed position) rather than horizon-scaled autoregressive waypoints, EdgeVTP reduces decoding overhead while producing smooth trajectories. To keep runtime predictable in crowded scenes, we explicitly bound interaction complexity via a locality graph with a hard neighbor cap. Across three highway benchmarks and two Jetson-class platforms, EdgeVTP achieves the lowest measured end-to-end latency under a protocol that includes graph construction and post-processing, while attaining state-of-the-art (SotA) prediction accuracy on two of the three datasets and competitive error on other benchmarks. Our code is available at \url{https://github.com/SeungjinStevenKim/EdgeVTP}.
\end{abstract}
\vspace{-10pt}

\section{Introduction}

Roadside camera networks are increasingly used for real-time highway safety monitoring, workzone safety, incident detection, and traffic management \cite{giannakeris2018speed,wei2018unsupervised,verma2024etram,hasanujjaman2023sensor,qiu2024intelligent,Ozer_2014_CVPR_Workshops,naphade2019aicity,naphade2020aicity, tang2019cityflow}. A core capability in these systems is vehicle trajectory prediction (VTP): forecasting the future motion of many interacting vehicles from noisy tracked observations \cite{xu2024adapting,zhang2024oostraj,tang2024hpnet,10657975,pourkeshavarz2024cadet,li2019grip,liao2023bat,wen2020uadetrac}. Compared to urban driving, highway scenes exhibit long-range, high-speed interactions (lane changes, merges, and stop-and-go waves) and can encompass up to hundreds of interacting agents in a single field of view, making multi-agent prediction both accuracy-critical and computationally demanding \cite{grimm2025goal, li2014stop, krajewski2018highd}. These challenges are amplified in surveillance-based settings, where fixed-camera viewpoints introduce perspective distortion and tracking noise, and where inference must often run continuously on near-roadway edge platforms with tight power and memory budgets \cite{singh2023edgeai,swaminathan2024jetsonnano}.

Prior work has made substantial progress on highway motion forecasting. Early approaches relied on recurrent encoder--decoder designs and explicit maneuver modeling to capture multi-modality \cite{lee2017desire}. Maneuver-based LSTMs and convolutional social pooling achieved strong freeway performance by combining temporal modeling with interaction-aware pooling and multi-modal decoding \cite{deo2018csp,deo2018mm, lin2021stalstm, alahi2016social, mohamed2020socialstgcnn}. More recent methods increasingly use structured interaction modeling and attention mechanisms \cite{li2019grip, chen2022vehicle}. Graph neural networks and map-centric encoders model actor-actor and actor--map relationships for improved accuracy \cite{gajiyang2020vectornet,liang2020lanegcn}, and probabilistic graph-structured predictors enable flexible multi-agent forecasting with heterogeneous inputs \cite{salzmann2020trajectronpp}. Transformer-based models further improve long-range temporal reasoning, including highway-surveillance benchmarks such as CHD with high-angle and eye-level viewpoints \cite{pazho2024vtformer,chd,shi2022mtr,nayakanti2022wayformer,zhang2023hptr}.

Despite these advances, embedded deployment introduces requirements that are often under-emphasized in accuracy-driven studies: (i) predictable end-to-end (E2E) runtime under dense traffic, (ii) tight memory and power budgets, and (iii) latency constraints for streaming video analytics and on-device decision loops \cite{singh2023edgeai,swaminathan2024jetsonnano,Apewokin_RealTimeAdaptiveBackground_2011,reddi2020mlperf,chen2018tvm,nvidia_tensorrt,jeong2022jedi}. In practice, dense scenes can increase end-to-end latency as interaction graphs grow, even when model-only runtime appears acceptable \cite{alinezhad2023pishgu}. A second bottleneck is decoding: autoregressive predictors incur inference cost that scales with the prediction horizon and can accumulate error when predictions are fed back into the model, motivating non-autoregressive alternatives that generate future sequences in parallel \cite{qi2020nart,achaji2022pretr}.

We present EdgeVTP, an embedded-first redesign of a transformer-graph predictor for surveillance-based highway trajectory prediction. EdgeVTP is designed to achieve a favorable trade-off between prediction accuracy and inference latency by making three deployment-driven principles explicit: one-shot parametric trajectory generation, bounded interaction complexity, and lightweight model design. First, we replace step-wise autoregressive decoding with a one-shot parametric head that regresses to a low-dimensional manifold of control points and reconstructs dense future trajectories via analytic evaluation, reducing sequential overhead while producing smooth trajectories by construction. Second, we sparsify the interaction graph using physically motivated neighborhood construction (radius gating and top-$K$ capping) to upper-bound interaction complexity and stabilize runtime as scene density increases. Third, we keep temporal modeling lightweight and deployment-oriented by using a compact temporal encoder and systematically slimming the transformer (layers/heads/width) to reduce compute and memory without sacrificing the ability to reason over multi-agent interactions.

Experiments across surveillance highway benchmarks and embedded platforms show that explicitly designing for bounded interaction cost and one-shot decoding yields a consistently better accuracy-latency trade-off under realistic E2E measurement, including graph construction and trajectory reconstruction. A key observation motivating our design is that, on 
Jetson-class edge devices, the dominant latency bottlenecks 
in transformer-graph VTP pipelines are not the learned 
parameters but rather (i)~scene-dependent graph construction, 
whose cost grows with traffic density, and 
(ii)~autoregressive decoding, whose cost scales with the 
prediction horizon. Standard model compression 
(pruning, distillation, quantization) does not address 
either bottleneck directly. EdgeVTP instead targets these 
two costs at the architecture level.

\noindent\textbf{Contributions.}
\begin{itemize}[leftmargin=*,noitemsep,topsep=0pt]
    \item We introduce EdgeVTP, an embedded-first Surveillance-based VTP (SVTP) architecture that frames roadside trajectory prediction as an end-to-end accuracy--latency co-design problem, making deployment constraints explicit through bounded interaction graphs and one-shot curve decoding.
    \item We expose interpretable operating-point knobs $(r, K,\text{ Residual})$ that span the accuracy--latency spectrum, and show that the preferred operating point depends on surveillance viewpoint (CHD Eye-level vs.\ High-angle).
    \item We establish an end-to-end latency protocol that includes graph construction and trajectory reconstruction, and demonstrate consistent deployability gains: lowest E2E latency on Jetson-class hardware among the compared methods while achieving State-of-the-Art (CHD) or competitive (NGSIM) prediction error.
\end{itemize}

\section{Related Work}
\label{sec:related}

Vehicle trajectory prediction (VTP) has evolved from classical kinematic motion models to deep architectures that model multi-agent interactions and long-range temporal dependencies. In embedded vision deployments, however, the dominant failure mode is often not average accuracy but unpredictable E2E runtime under scene density shifts (e.g., rush-hour traffic), where interaction graphs become large and decoding cost scales with the prediction horizon. Our work targets this deployment gap by (i) enforcing bounded interaction complexity and (ii) eliminating autoregressive decoding via one-shot parametric curve prediction.

\textbf{Surveillance-based VTP and Real-time Highway Monitoring.}
Surveillance-based VTP (SVTP) differs from egocentric forecasting due to fixed-camera viewpoints, perspective distortion, and noisier tracking signals \cite{dubska2014trafficcalib,sochor2017trafficcalib,wen2020uadetrac,naphade2020aicity,tang2019cityflow}. In this setting, lightweight temporal modeling has been emphasized for real-time roadway monitoring. DeepTrack introduced an embedded-friendly temporal convolutional design for highway prediction and monitoring under constrained compute budgets \cite{katariya2022deeptrack}. More recently, VT-Former explored combining interaction modeling with transformer-based decoding for highway surveillance prediction \cite{ngiam2021scene, yuan2021agentformer, pazho2024vtformer}. These works motivate SVTP-specific designs that remain robust to viewpoint effects while delivering predictable runtime under the practical constraints of roadside deployments. Highway benchmarks such as  NGSIM \cite{NGSIM_i80, NGSIM_US101}, CHD\cite{chd}, HighD \cite{krajewski2018highd}, exiD\cite{moers2022exid} are used widely for SVTP application. Autonomous driving benchmarks such as nuScenes \cite{caesar2020nuscenes}, Waymo \cite{sun2020scalability}, Argoverse \cite{chang2019argoverse, wilson2023argoverse}, Zenseact open dataset \cite{alibeigi2023zenseact} also used for VTP but are rarely used in fixed-camera surveillance settings, aside from scenarios such as knowledge distillation.

\textbf{Interaction Modeling: Social Pooling to Graph Neural Networks.}
Capturing inter-dependencies between vehicles is essential for accurate VTP in dense highway traffic. Convolutional Social Pooling (CSP) pioneered grid-based interaction encoding via a convolutional pooling layer over surrounding vehicles \cite{deo2018csp}. Subsequent work shifted toward graph neural networks (GNNs), which represent agents as nodes and interactions as edges, enabling structured message passing on non-Euclidean relationships. GRIP demonstrated graph-based interaction-aware trajectory prediction in traffic scenarios \cite{li2019grip}, while Social-STGCNN showed that spatio-temporal graph convolutions can provide interaction modeling with strong efficiency characteristics \cite{mohamed2020socialstgcnn,gupta2018social,huang2025trajectory,zhang2023hptr}. For edge-centric cyber-physical systems, Pishgu further studied GIN-style interaction modeling and reported real-time performance on embedded platforms \cite{alinezhad2023pishgu}. Graph attention mechanisms (e.g., GAT) provide adaptive neighbor weighting, improving expressiveness when interaction strength varies across neighbors \cite{velickovic2018gat}.

While these approaches improve interaction awareness, many formulations implicitly allow interaction cost to grow with scene density (e.g., fully-connected or high-degree graphs). This is a key embedded challenge: dense scenes can substantially degrade E2E latency in real-time monitoring. On edge devices, this often shows up as latency increasing sharply with agent count, even when average-case runtime looks acceptable \cite{alinezhad2023pishgu}. Our work directly addresses this scaling issue by bounding neighborhood size (via geometric filtering and hard caps) to upper-bound interaction complexity.

\textbf{Spatio-Temporal Attention and Lightweight Sequence Modeling.}
Attention mechanisms have been used to focus prediction on the most relevant historical states and neighboring agents. STA-LSTM employs spatial and temporal attention to quantify the influence of specific neighbors and past steps, providing both accuracy gains and interpretability via attention weights \cite{alahi2016social, lin2021stalstm}. In parallel, SVTP-oriented hybrid designs (e.g., VT-Former) leverage interaction-aware tokenization and transformer components to model longer-range dependencies \cite{vaswani2017attention, pazho2024vtformer, zhou2022hivt}. For embedded deployment, these approaches highlight the need to retain the benefits of attention while controlling compute growth under dense traffic. Our design follows the same principle, preserving interaction reasoning, while making runtime more predictable through explicit bounds on the interaction graph. Recent work has also explored lightweight transformer designs for trajectory forecasting with reduced overhead, e.g., hierarchical light transformer ensembles \cite{lafage2025hltens}.

\textbf{Fast Decoding and Non-autoregressive Trajectory Generation.}
A second embedded bottleneck is decoding: many predictors generate future steps sequentially, leading to inference cost that scales with the prediction horizon and accumulating error when predictions are fed back into the model. Recent work in trajectory prediction has explored non-autoregressive decoding strategies that generate futures in parallel \cite{phanminh2020covernet,chai2020multipath,zhao2020tnt,gu2021densetnt}. For example, TUTR unifies social interaction and multimodal trajectory prediction in a transformer encoder–decoder design and predicts diverse trajectories in parallel, demonstrating that parallel decoding can reduce test-time overhead compared to pipelines that rely on expensive post-processing \cite{gupta2018social, Shi_2023_ICCV}. PreTR similarly leverages parallel decoding with learned queries to mitigate exposure bias and reduce test-time computation \cite{achaji2022pretr}.

Complementary to non-autoregressive decoding, compact parametric decoding reduces step-wise cost by predicting a small parameter set and reconstructing dense trajectories. Probabilistic B\'ezier curve formulations provide a representative control-point-based approach for one-shot multi-step prediction \cite{hug2020probbezier}. In the autonomous-driving setting, Efficient Motion Prediction (EMP) further illustrates that carefully chosen lightweight architectural components can achieve strong accuracy with fast training and inference \cite{prutsch2024efficient}. Our work builds on this line of thinking for SVTP: we remove iterative decoding entirely by predicting a small set of curve parameters in one pass, enabling predictable runtime on edge hardware.

\textbf{Behavior-aware Forecasting under Embedded Constraints.}
A recent direction is to incorporate cognitive priors and behavioral insights to better handle intent uncertainty. HLTP adopts a teacher--student framework inspired by human visual processing and visual attention allocation \cite{liao2024hltp}. BAT introduces a behavior-aware model that integrates traffic-psychology motivated cues to infer interactions without manual behavior labeling \cite{liao2023bat}. These approaches can improve accuracy and robustness, but their additional behavioral reasoning components and training pipelines can increase system complexity. In contrast, our focus is on deployment-facing efficiency: bounding interaction cost and simplifying decoding, while remaining compatible with future integration of richer behavioral priors.

\textbf{Embedded Efficiency and Deployment-oriented Evaluation.}
Embedded VTP systems often rely on model compression and hardware-aware optimization (e.g., distillation, pruning, and quantization) to reduce compute and memory footprints. Knowledge distillation (KD) is a widely used paradigm for transferring knowledge from a larger teacher model to a compact student network \cite{hinton2015distilling, gou2021kd}. In embedded vision practice, latency-aware structured pruning has been explicitly studied to align sparsity with measured execution cost on target devices \cite{Belhadi_2025_ICCV, Chandorkar_2025_ICCV, Cigla_2018_CVPR_Workshops} . Hardware/algorithm co-design for low-bit inference can further reduce compute, as illustrated by EVW work on binary-weight networks and dedicated inference engines \cite{Chen_2024_CVPR}.

These techniques are complementary to our approach. In many real-time SVTP deployments, worst-case latency is frequently dominated by (i) scene-dependent interaction density and (ii) decoding cost rather than raw parameter count alone. Accordingly, we focus on architecture-level predictability (bounded graphs and one-shot decoding), while treating compression methods as orthogonal improvements. Finally, EVW work on low-latency embedded vision emphasizes that deployment claims should be supported by E2E measurements rather than forward-pass timing alone \cite{Prabhune_2024_CVPR}. Our evaluation follows this deployment-first perspective by timing the full inference pipeline on edge hardware.

\section{Method}
\label{sec:method}

We present EdgeVTP, an embedded-first model for vehicle trajectory prediction in roadside highway surveillance. EdgeVTP is built to deliver a favorable accuracy-latency trade-off under E2E measurement, where runtime includes interaction graph construction and trajectory reconstruction. The design is guided by three principles: (i) one-shot future generation to avoid horizon-scaled autoregressive decoding, (ii) bounded interaction complexity to stabilize runtime in dense scenes, and (iii) compact model design to reduce compute and memory for edge deployment.

\begin{figure*}[t]
  \centering
  \includegraphics[width=\textwidth]{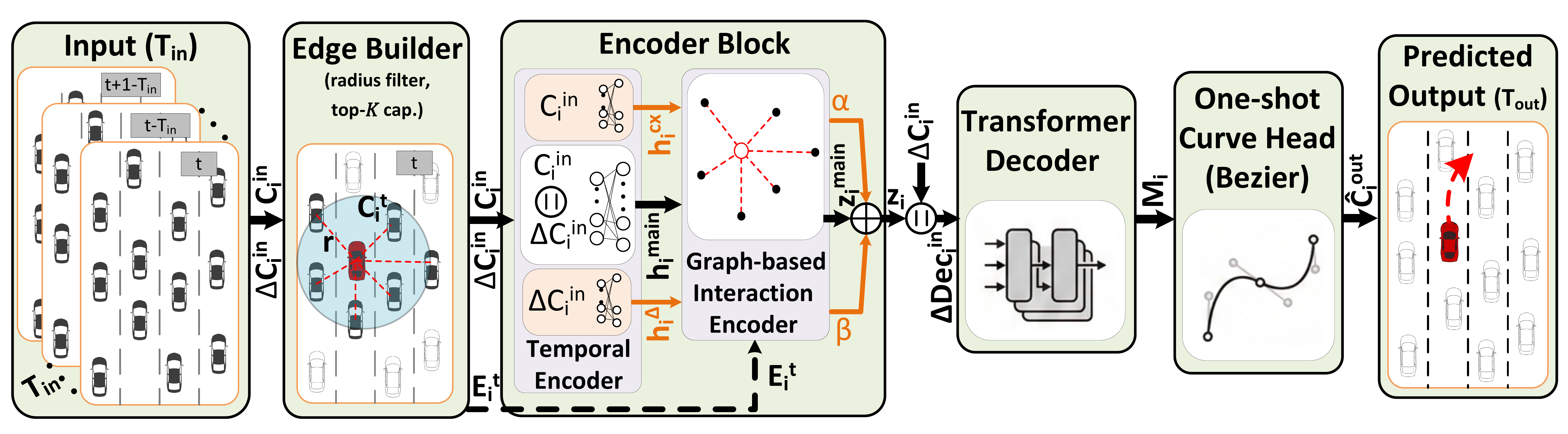}
  \caption{\textbf{EdgeVTP overview.} For each vehicle, we use observed absolute positions $\mathbf{C}_i^{\text{in}}(t)$ and displacements $\Delta\mathbf{C}_i^{\text{in}}(t)$ over $T_{\text{in}}$ frames. The Edge Builder forms a directed neighbor graph at time $t$ using a radius $r$ and top-$K$ cap, producing edge indices $E_i^{t}$. A temporal encoder projects the motion history, and a Graph-based interaction encoder (GIE) aggregates neighbor information; branch features are fused via optional learnable scalar residual weighting ($\alpha,\beta$) to produce $\mathbf{z}_i$ (Optional branches are shown as orange arrows). To preserve local motion fidelity while incorporating social context, we fuse interaction embeddings with raw displacement histories as decoder memory tokens ($\mathbf{Dec}_i^{\text{in}}(t)=[\Delta\mathbf{C}_i^{\text{in}}(t)\;;\;\mathbf{Z}_i(t)]$). Then a compact Transformer decoder drives a one-shot degree-4 B\'ezier head that predicts four control point offsets $\{\Delta P_k\}_{k=1}^{4}$ to define a 5-point degree-4 curve anchored at $P_0$ and samples $T_{\text{out}}$ future positions to output $\widehat{\mathbf{C}}_i^{\text{out}}(t)$.}
  \label{fig:edgevtp_arch}
\end{figure*}

\subsection{Problem Setup and Notation}
\label{sec:setup}

We consider a traffic scene with $N$ tracked vehicles at the current (last observed) time $t$. The 2D position of vehicle $i$ at time $\tau$ is $\mathbf{c}_i^{\tau}\in\mathbb{R}^2$. For each vehicle, we use an observed history of length $T_{\text{in}}$,
$\mathbf{C}_i^{\text{in}}(t)\in\mathbb{R}^{T_{\text{in}}\times 2}$, containing positions from $t-T_{\text{in}}+1$ to $t$.
We also use per-step displacements $\Delta\mathbf{c}_i^{\tau}=\mathbf{c}_i^{\tau}-\mathbf{c}_i^{\tau-1}$, forming a temporal displacement history
$\Delta\mathbf{C}_i^{\text{in}}(t)\in\mathbb{R}^{T_{\text{in}}\times 2}$ (with the first term padded for shape consistency).

Given these observations, the goal is to predict the future trajectory over a horizon $T_{\text{out}}$,
$\widehat{\mathbf{C}}_i^{\text{out}}(t)\in\mathbb{R}^{T_{\text{out}}\times 2}$, corresponding to positions from $t+1$ to $t+T_{\text{out}}$.
We denote the ground-truth future as $\mathbf{C}_i^{\text{out}}(t)$ and train the model to minimize prediction error over the forecast horizon.

\subsection{Edge Builder}
\label{sec:edge_builder}

As shown in Fig.~\ref{fig:edgevtp_arch}, the Edge Builder constructs an interaction graph at the current frame $t$ using only the last observed positions $\{\mathbf{c}_i^{t}\}_{i=1}^{N}$. For each vehicle $i$, we first form a candidate neighbor set by radius gating: we retain vehicles whose Euclidean distance to $i$ is at most $r$ meters. This removes distant actors that are unlikely to affect short-horizon motion and reduces graph density.

To make runtime predictable on embedded hardware, we then apply a hard top-$K$ cap. Among the radius candidates, we keep only the $K$ closest neighbors for each vehicle. We denote the resulting (capped) neighbor index list for vehicle $i$ as $E_i^{t}$, i.e., $E_i^{t}$ contains the indices of vehicle that exchange messages with $i$ at time $t$. This explicitly bounds the number of incoming interaction edges per vehicle, stabilizing message passing cost under varying traffic density and providing a direct efficiency knob: smaller $K$ reduces edge count, intermediate activations, and memory bandwidth pressure.

Aggregating all per-vehicle neighbor lists gives a directed edge set $\mathcal{E}(t)$, consisting of pairs $(i,j)$ for all $j\in E_i^{t}$. Neighbor selection is deterministic at inference time. During training, when more than K candidates exist within radius r during training, we sample K neighbors uniformly at random from the radius-filtered set; inference always uses the deterministic K-nearest selection. We ablate the effect of the interaction radius $r$ and neighbor cap $K$ on the accuracy--latency trade-off in the supplementary material.

\subsection{Encoder}
\label{sec:encoder}

Fig.~\ref{fig:edgevtp_arch} summarizes our encoder, which has two stages: a temporal projection that maps each vehicle's $T_{\text{in}}$ observed history to a compact embedding, followed by a graph-based interaction encoder that aggregates neighbor information using the edge index from the Edge Builder.

\noindent\textbf{Temporal encoding.}
For each vehicle $i$, we use two input streams (Sec.~\ref{sec:setup}): absolute positions $\mathbf{C}_i^{\text{in}}(t)\in\mathbb{R}^{T_{\text{in}}\times 2}$ and displacements $\Delta\mathbf{C}_i^{\text{in}}(t)\in\mathbb{R}^{T_{\text{in}}\times 2}$.Figure~\ref{fig:edgevtp_arch} depicts these histories as $T_{\text{in}}\times 2$ sequences; for the temporal encoder we flatten each stream and concatenate them into a single vector:
:$\mathbf{x}_i^{\mathrm{cx}}=\mathrm{vec}(\mathbf{C}_i^{\text{in}}(t))\in\mathbb{R}^{2T_{\text{in}}}$,
$\mathbf{x}_i^{\Delta}=\mathrm{vec}(\Delta\mathbf{C}_i^{\text{in}}(t))\in\mathbb{R}^{2T_{\text{in}}}$, and
$\mathbf{x}_i^{\mathrm{real}}=[\mathbf{x}_i^{\mathrm{cx}};\mathbf{x}_i^{\Delta}]\in\mathbb{R}^{4T_{\text{in}}}$.
The temporal encoder block applies a lightweight feed-forward projection to obtain an intermediate node embedding
$\mathbf{h}_i^{\mathrm{main}}=f_{\mathrm{temp}}(\mathbf{x}_i^{\mathrm{real}})$, implemented as a compact linear layer followed by LeakyReLU for efficiency.
To support residual separation (the two side branches in Fig.~\ref{fig:edgevtp_arch}), we optionally compute stream-specific embeddings
$\mathbf{h}_i^{\mathrm{cx}}=f_{\mathrm{temp}}^{\mathrm{cx}}(\mathbf{x}_i^{\mathrm{cx}})$ and
$\mathbf{h}_i^{\Delta}=f_{\mathrm{temp}}^{\Delta}(\mathbf{x}_i^{\Delta})$,
using the same lightweight projection structure.

\noindent\textbf{Graph interaction encoding (GIE).}
Given the per-vehicle neighbor index list $E_i^{t}$ (Sec.~\ref{sec:edge_builder}), we apply graph message passing to aggregate information from interacting vehicles. We compute a main interaction feature $\mathbf{z}_i^{\mathrm{main}}$ by applying a GIE block to $\mathbf{h}_i^{\mathrm{main}}$ over the directed edges defined by $E_i^{t}$.
When residual separation is enabled, we additionally apply separate GIE blocks to the stream-specific embeddings to obtain branch interaction features $\mathbf{z}_i^{\mathrm{cx}}$ and $\mathbf{z}_i^{\Delta}$. We align these branch features to the main dimensionality using a lightweight linear projection $\mathrm{Proj}(\cdot)$ and fuse them with learnable gates $\alpha$ and $\beta (\,\text{Fig.}~\ref{fig:edgevtp_arch}\,)$:
\begin{equation}
\mathbf{z}_i
=
\mathbf{z}_i^{\mathrm{main}}
+
\alpha\,\mathrm{Proj}(\mathbf{z}_i^{\mathrm{cx}})
+
\beta\,\mathrm{Proj}(\mathbf{z}_i^{\Delta}).
\label{eq:residual_fusion}
\end{equation}
The fused interaction feature $\mathbf{z}_i$ is then converted to the sequence representation used by the Transformer decoder and combined with the displacement stream to form the decoder memory tokens. Here, $\alpha$ and $\beta$ are learnable scalar residual weights (shared across vehicles and time) that weight the two residual branches; we initialize $\alpha=\beta=0.1$ and learn them end-to-end so that the residual branches contribute modestly at the start of training. These weights allow the model to dynamically balance the influence of absolute position vs. displacement streams.

\subsection{Transformer Decoder}
\label{sec:decoder}

\noindent\textbf{Memory construction.}
The encoder output (Eq.~\ref{eq:residual_fusion}) is converted to a sequence aligned with the input horizon and denoted as $\mathbf{Z}_i(t)$ (Fig.~\ref{fig:edgevtp_arch}). We then concatenate $\mathbf{Z}_i(t)$ with the observed displacement history $\Delta\mathbf{C}_i^{\text{in}}(t)$ along the feature dimension to form the decoder input tokens:
$\mathbf{Dec}_i^{\text{in}}(t)=[\Delta\mathbf{C}_i^{\text{in}}(t)\;;\;\mathbf{Z}_i(t)]$.
This $T_{\text{in}}$-length sequence $\mathbf{Dec}_i^{\text{in}}(t)$ is passed to the Transformer decoder.

\noindent\textbf{Lightweight decoding.}
We use a compact Transformer decoder (2 layers, 2 heads) that attends to $\mathbf{Dec}_i^{\text{in}}(t)$ and produces a single latent token $\mathbf{M}_i(t)$ for prediction. We ablate the effect of decoder capacity (e.g., number of attention heads) on accuracy and latency in the supplementary material. Importantly, the decoder does \emph{not} generate future steps autoregressively; instead, $\mathbf{M}_i(t)$ parameterizes the one-shot degree-4 B\'ezier head described in Sec.~\ref{sec:bezier}.




\subsection{One-shot Degree-4 B\'ezier Decoding}
\label{sec:bezier}

We replace horizon-scaled autoregressive decoding with a one-shot, curve-parameterized prediction head. For each vehicle $i$, we anchor the curve at the last observed position $\mathbf{P}_0=\mathbf{c}_i^{t}$. The Transformer decoder produces a single latent token $\mathbf{M}_i(t)$ (from previous Sec.~\ref{sec:decoder}), which a lightweight linear head maps to four 2D control point offsets. These, combined with the anchor $P_0$, define the set $\{P_0, \dots, P_4\}$, defining the remaining control points of a degree-4 B\'ezier curve:
\begin{equation}
\mathbf{P}_k = \mathbf{P}_0 + \Delta\mathbf{P}_k,\qquad k\in\{1,2,3,4\}.
\label{eq:bezier_ctrlpts}
\end{equation}
Given the five control points $\{\mathbf{P}_0,\dots,\mathbf{P}_4\}$, we obtain the predicted future trajectory by evaluating the B\'ezier curve at uniformly spaced parameters $u_s\in[0,1]$:
\begin{equation}
\widehat{\mathbf{c}}_i^{\,t+s}
=
\sum_{k=0}^{4} \binom{4}{k}(1-u_s)^{4-k}u_s^{k}\,\mathbf{P}_k,
\qquad s=1,\dots,T_{\text{out}},
\label{eq:bezier_curve}
\end{equation}
where u$_s$ = s/T$_{out}$. Unless stated otherwise, we use $T_{\text{out}}=25$.

This formulation predicts a small set of control points rather than $T_{\text{out}}$ step-wise outputs, eliminating sequential decoding and reducing output-layer computation and activation memory. At the same time, the global curve parameterization encourages temporally coherent (smooth) forecasts while remaining flexible enough to represent common highway maneuvers.

\subsection{Training Objective}
\label{sec:loss}

The one-shot B\'ezier head (Sec.~\ref{sec:bezier}) yields $T_{\text{out}}=25$ future positions $\{\widehat{\mathbf{c}}_{i}^{t+k}\}_{k=1}^{T_{\text{out}}}$ for each vehicle. We train using a masked $\ell_2$ loss in absolute space (mask $m_i^k$ accounts for vehicle visibility), normalized by the total number of prediction slots $N\cdot T_{\text{out}}$.
\begin{equation}
\mathcal{L}_{\text{traj}}
=\frac{1}{N\cdot T_{\text{out}}}
\sum_{i=1}^{N}\sum_{k=1}^{T_{\text{out}}}
m_i^{k}\,\big\|\widehat{\mathbf{c}}_{i}^{t+k}-\mathbf{c}_{i}^{t+k}\big\|_2^{2}.
\end{equation}

\noindent\textbf{Design for embedded deployment.}
The EdgeVTP architecture is explicitly designed to maintain low and predictable end-to-end latency in dense traffic scenarios. Bounded interaction graphs (radius gating with an optional top-$K$ cap) limit message passing cost under dense traffic, while a compact decoder keeps attention and activation memory small. Finally, one-shot B\'ezier decoding removes horizon-scaled autoregressive steps in favor of a single decoder pass followed by analytic curve evaluation. Together, these choices reduce compute and memory footprints while preserving interaction-aware forecasting.

\section{Experimental Setup}
\label{sec:exp_setup}

\noindent\textbf{Datasets and protocol.}
We evaluate on three surveillance highway settings: NGSIM and CHD under two camera viewpoints (High-angle and Eye-level). NGSIM features dense highway interactions, while CHD complements it with viewpoint and scene-geometry variation that stresses robustness to perspective distortion and tracking noise.
Each sample consists of an observed history of $T_{\text{in}}=15$ steps and a prediction horizon of $T_{\text{out}}=25$ steps. To ensure a consistent temporal window across datasets, we resample sequences to a common rate (5\,Hz), corresponding to 3\,s of observation and 5\,s of prediction. We follow the standard dataset splits and evaluation protocol used in prior work for each benchmark.

\noindent\textbf{Model configuration and operating points.}
Unless stated otherwise, EdgeVTP uses a lightweight temporal projection (single FC with LeakyReLU) followed by a GIN-based interaction encoder, a compact Transformer decoder (2 layers, 2 heads), and one-shot degree-4 B\'ezier decoding that predicts four control-point offsets and analytically reconstructs $T_{\text{out}}$ future positions (Sec.~\ref{sec:bezier}). For graph construction (Sec.~\ref{sec:edge_builder}), we apply radius gating with a hard top-$K$ neighbor cap; during training, we optionally random-sample within the radius as regularization while preserving the same test-time compute budget.
From a sweep over $(r, K)$ and residual separation, we select three representative operating points (Table \ref{tab:ngsim_3models} obtained from ablation study presented in the supplementary material) that we report consistently across the paper; the complete sweep is provided in the supplementary material.

\noindent\textbf{Training details and hardware.}
Unless stated otherwise, training and evaluation are performed on NVIDIA H100 GPUs. We train for 80 epochs using Adam with learning rate $1\times10^{-2}$ and weight decay $5\times10^{-4}$. We use a multi-step schedule with milestones at epochs 40/60/70 and decay factor 0.1. The batch size is 16 and dropout is 0.2. Our primary objective is a masked $\ell_2$ loss in absolute position space (Sec.~\ref{sec:loss}).

\noindent\textbf{Metrics and latency measurement.}
We report ADE (average displacement error) and FDE (final displacement error), along with horizon-specific RMSE at 1-5 seconds (computed at 5\,Hz). Errors are reported in meters on NGSIM and in pixels on CHD, following each dataset protocol.
We measure latency with batch size 1 and report E2E runtime, which includes graph construction (Edge Builder), model inference, and post-processing (trajectory reconstruction). We run a warm-up phase (50 runs) and report mean latency over repeated inference on the NGSIM test set under a fixed deployment configuration. Latency is measured using PyTorch FP32 inference with synchronization for timing. To reduce measurement noise, we keep the software stack and precision fixed, disable data loader workers during timing (\texttt{num\_workers=0}), fix CPU threading, and avoid logging or metric computation inside the timed region. We use H100 timings to analyze relative compute scaling under a controlled stack, and we report embedded-device latency separately on Jetson platforms.

\section{Results}
\label{sec:results}

We evaluate EdgeVTP under two coupled objectives central to embedded deployment: prediction quality and deployability. We report prediction errors (ADE/FDE and horizon RMSE), parameter count, and E2E latency (batch size 1) on NGSIM (meters) and CHD under two surveillance viewpoints (pixels), following each dataset protocol. 

\begin{table}[!h]
\centering
\caption{CHD Eye-level: comparison to established baselines on the CHD benchmark \cite{chd}. Errors in pixels (lower is better).}
\label{tab:chd_e3}
\resizebox{\columnwidth}{!}{%
\begin{tabular}{@{}ccc|ccccc@{}}
\toprule
\multicolumn{1}{l}{\textbf{}} & \multicolumn{1}{l}{\textbf{}} & \multicolumn{1}{l}{\textbf{}} & \multicolumn{5}{|c}{\textbf{RMSE}}                                   \\ \midrule
\textbf{Model}                & \textbf{ADE}                  & \textbf{FDE}                  & \textbf{1s} & \textbf{2s} & \textbf{3s} & \textbf{4s} & \textbf{5s} \\\midrule
S-STGCNN \cite{mohamed2020socialstgcnn}     & 24.33 & 95.22  & 4.32 & 9.15  & 15.93 & 29.05 & 68.32 \\
GRIP++ \cite{li2019grip++}              & 44.27 & 129.58 & 4.42 & 12.86 & 24.31 & 35.04 & 145.17 \\
Pishgu \cite{alinezhad2023pishgu}       & 37.99 & 123.69 & 4.98 & 13.58 & 26.61 & 50.31 & 106.45 \\
VT-Former$_{LH}$ \cite{pazho2024vtformer}                     & 34.88 & 100.59 & 6.71 & 17.24 & 28.70 & 45.86 & 82.00 \\
VT-Former$_{MH}$  \cite{pazho2024vtformer}                      & 27.44 & 85.45  & 5.19 & 12.90 & 21.38 & 35.09 & 68.60 \\
VT-Former$_{SH}$  \cite{pazho2024vtformer}                       & 21.86 & 66.28  & 5.42 & 12.69 & 18.81 & 26.67 & 53.05 \\
\midrule
\rowcolor[HTML]{D9D9D9}
EdgeVTP$_{\text{Lat}}$           & 27.53 & 84.62  & 7.50 & 15.90 & 27.78 & 48.82 & 94.64 \\
\rowcolor[HTML]{D9D9D9}
EdgeVTP$_{\text{TF}}$            & 19.24 & 56.55  & 6.21 & 12.35 & 19.59 & 31.26 & 60.79 \\
\rowcolor[HTML]{D9D9D9}
EdgeVTP$_{\text{Error}}$          & 29.65 & 98.04  & 7.64 & 16.07 & 28.86 & 54.20 & 112.25 \\
\bottomrule
\end{tabular}%
}
\end{table}

\noindent\textbf{CHD: robustness across viewpoints.}
We report three EdgeVTP operating points: EdgeVTP$_{\text{Lat}}$ (latency-focused), EdgeVTP$_{\text{TF}}$ (balanced), and EdgeVTP$_{\text{Error}}$ (accuracy-focused), selected via an NGSIM ablation sweep.

Tables~\ref{tab:chd_e3} and \ref{tab:chd_h3} report results on CHD Eye-level and High-angle splits, respectively (pixels). A key observation is that the preferred operating point depends on viewpoint: the balanced configuration performs best on Eye-level scenes, while the accuracy-focused configuration performs best under the overhead High-angle viewpoint. Across both splits, EdgeVTP improves over other transformer-based baselines and remains competitive with the strongest published methods, indicating that the embedded-oriented design transfers beyond the dataset used for operating-point selection. For CHD, we apply the same radius gating in the dataset coordinate system (i.e., r is in pixels).


\begin{table}[!h]
\centering

\caption{CHD High-angle: comparison to established baselines on the CHD benchmark \cite{chd}. Errors in pixels (lower is better).}
\label{tab:chd_h3}
\resizebox{\columnwidth}{!}{%
\begin{tabular}{@{}ccc|ccccc@{}}
\toprule
\multicolumn{1}{l}{}   & \multicolumn{1}{l}{} & \multicolumn{1}{l}{} & \multicolumn{5}{|c}{\textbf{RMSE}}                                   \\ \midrule
\textbf{Model}         & \textbf{ADE}         & \textbf{FDE}         & \textbf{1s} & \textbf{2s} & \textbf{3s} & \textbf{4s} & \textbf{5s} \\
\midrule
S-STGCNN \cite{mohamed2020socialstgcnn} & 31.87 & 98.46  & 9.74 & 21.83 & 29.01 & 42.34 & 82.14 \\
GRIP++ \cite{li2019grip++}        & 36.32 & 100.89 & 3.40 & 6.67  & 14.32 & 28.02 & 123.04 \\
Pishgu \cite{alinezhad2023pishgu} & 18.33 & 61.92  & 4.04 & 7.48  & 13.99 & 24.30 & 51.51 \\
VT-Former$_{LH}$ \cite{pazho2024vtformer}                 & 25.95 & 87.21  & 7.60 & 17.35 & 22.90 & 27.97 & 66.39 \\
VT-Former$_{MH}$  \cite{pazho2024vtformer}              & 25.90 & 87.90  & 6.40 & 14.62 & 20.56 & 29.44 & 70.05 \\
VT-Former$_{SH}$ \cite{pazho2024vtformer}               & 25.33 & 88.99  & 5.67 & 12.96 & 19.12 & 29.83 & 70.72 \\
\midrule
\rowcolor[HTML]{D9D9D9}
EdgeVTP$_{\text{Lat}}$    & 22.04 & 74.04  & 5.77 & 12.36 & 21.42 & 37.41 & 82.83 \\
\rowcolor[HTML]{D9D9D9}
EdgeVTP$_{\text{TF}}$     & 18.77 & 60.04  & 5.61 & 11.48 & 19.10 & 31.99 & 68.14 \\
\rowcolor[HTML]{D9D9D9}
EdgeVTP$_{\text{Error}}$  & 15.23 & 52.28  & 4.38 & 8.71  & 14.55 & 26.01 & 59.50 \\
\bottomrule
\end{tabular}%
}
\end{table}

\begin{table}[b]
\centering
\caption{Ablation-selected EdgeVTP operating points on NGSIM (meters). Full ablation sweep is reported in the supplementary material.}
\label{tab:ngsim_3models}
\resizebox{\columnwidth}{!}{%
\begin{tabular}{lcccccc}
\toprule
\textbf{Variant} & \textbf{$r$} & \textbf{$K$} & \textbf{Residual} & \textbf{ADE} & \textbf{FDE} & \textbf{E2E (ms)} \\
\midrule
EdgeVTP$_{\text{Lat}}$   & 20m & 16 & No  & 2.13 & 4.93 & \textbf{3.17} \\
EdgeVTP$_{\text{TF}}$    & 20m & 16 & Yes & 1.89 & 4.37 & 4.30 \\
EdgeVTP$_{\text{Error}}$ & 30m & 16 & Yes & \textbf{1.85} & 4.25 & 4.58 \\
\bottomrule
\end{tabular}%
}
\end{table}

\noindent\textbf{Ablation-guided operating points.}
EdgeVTP exposes explicit deployment knobs through the Edge Builder (interaction radius $r$ and neighbor cap $K$) and an optional residual refinement path. We sweep these choices on NGSIM as discussed previously in this section to select three representative operating points that we use consistently across all benchmarks. Table~\ref{tab:ngsim_3models} summarizes these variants. The full sweep is reported in the supplementary material and shows the resulting accuracy-latency trade-off. Our profiling reveals that pipeline overhead, including graph construction, accounts for 25-35\% of total E2E runtime on edge hardware, justifying our focus on architectural predictability.

\begin{table*}[!t]
\centering
\caption{Results comparison on the NGSIM datasets\cite{NGSIM_i80, NGSIM_US101}. Errors in meters; E2E latency is batch=1 when available (lower is better). Missing entries are denoted by \textbf{--}. $^*$ represents models with latencies obtained from literature.}
\label{tab:ngsim_main}
\begingroup
\footnotesize
\setlength{\tabcolsep}{4pt}%
\begin{tabular*}{0.97\textwidth}{@{\extracolsep{\fill}}lcc|ccccc|ccc@{}}
\toprule
\textbf{Model} & \textbf{ADE} & \textbf{FDE} &
\textbf{1s} & \textbf{2s} & \textbf{3s} & \textbf{4s} & \textbf{5s} &
\textbf{AVG} & \textbf{Params (K)} & \textbf{E2E (ms)} \\
\midrule
Pishgu$^*$ \cite{alinezhad2023pishgu}       & 2.44 & 5.39 & 0.60 & 1.77 & 3.09 & 4.55 & 6.15 & 3.232  & 132   & 3.50 \\
iNATran$^*$ \cite{chen2022vehicle}       & \textbf{--} & \textbf{--} & 0.39 & 0.96 & 1.61 & 2.42 & 3.43 & 1.762  & \textbf{--} & 20.92 \\
DTBP$^*$ \cite{gao2023dual}          & \textbf{--} & \textbf{--} & 1.18 & 2.83 & 4.22 & 5.82 & \textbf{--} & 3.5125 & \textbf{--} & 62.00 \\
CS-LSTM$^*$  \cite{deo2018csp}      & 2.29 & 3.34 & 0.61 & 1.27 & 2.09 & 3.10 & 4.37 & 2.288  & 191   & 3.61 \\
STA-LSTM  \cite{lin2021stalstm}     & 1.89 & 3.16 & 0.37 & 0.98 & 1.17 & 2.63 & 3.78 & 1.786  & 124   & 5.01 \\
DeepTrack \cite{katariya2022deeptrack}     & 2.01 & 3.21 & 0.47 & 1.08 & 1.83 & 2.75 & 3.89 & 2.004  & 109   & \textbf{--} \\
VT-Former$_{LH}$ \cite{pazho2024vtformer}   & 2.53 & 5.51 & 0.46 & 1.33 & 2.35 & 3.50 & 4.80 & 2.488  & 155   & 31.69 \\
VT-Former$_{MH}$\cite{pazho2024vtformer}    & 2.24 & 5.15 & 0.37 & 1.13 & 2.10 & 3.26 & 4.62 & 2.296  & 141   & 26.86 \\
VT-Former$_{SH}$ \cite{pazho2024vtformer}   & 2.10 & 4.91 & 0.30 & 0.99 & 1.90 & 3.00 & 4.31 & 2.10   & 132   & 23.69 \\
\midrule

\rowcolor[HTML]{D9D9D9}
EdgeVTP$_{\text{Lat}}$   & 2.13 & 4.93 & 0.59 & 1.40 & 2.42 & 3.63 & 5.01 & 2.61 & 134.5 & 3.17 \\
\rowcolor[HTML]{D9D9D9}
EdgeVTP$_{\text{TF}}$   & 1.89 & 4.37 & 0.56 & 1.26 & 2.15 & 3.24 & 4.52 & 2.35 & 145.9 & 4.30 \\
\rowcolor[HTML]{D9D9D9}
EdgeVTP$_{\text{Error}}$ & 1.85 & 4.25 & 0.60 & 1.31 & 2.19 & 3.25 & 4.51 & 2.37 & 145.9 & 4.58 \\
\bottomrule
\end{tabular*}%
\endgroup
\end{table*}

\begin{figure}[t]
  \centering
  \includegraphics[width=1\columnwidth,keepaspectratio]{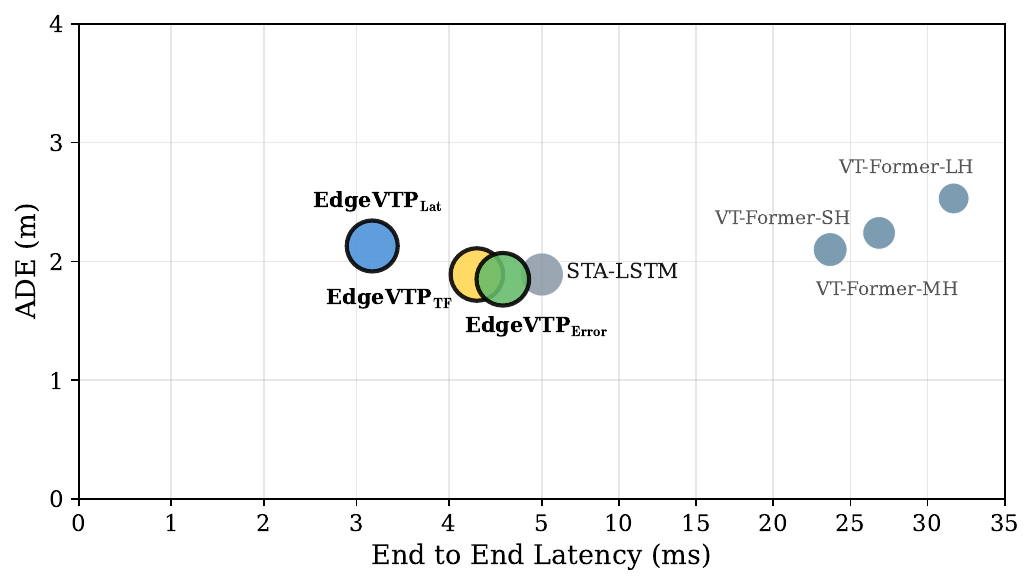}
    \caption{Accuracy–latency trade-off on NGSIM. Each point represents a model evaluated on the NGSIM dataset. The ideal operating region is the bottom-left (closer to origin is better). Highlighted points indicate the three operating points used in the main paper: EdgeVTP$_{\text{Lat}}$ (latency-focused), EdgeVTP$_{\text{TF}}$ (balanced), and EdgeVTP$_{\text{Error}}$ (accuracy-focused). All points report E2E latency under the protocol in Section  \ref{sec:exp_setup}}
  \label{fig:ade_e2e_scatter}
\end{figure}

\noindent\textbf{NGSIM: comparison to prior methods.}
Table~\ref{tab:ngsim_main} compares EdgeVTP to representative trajectory predictors on NGSIM (meters) while also reporting E2E latency when available. Across baselines, strong accuracy often comes with higher runtime, particularly for heavier transformer-centric pipelines. In contrast, EdgeVTP provides a practical accuracy-latency trade-off: the latency-focused operating point achieves the fastest E2E inference, while the balanced and accuracy-focused variants improve prediction quality with only modest additional runtime. These results support the central goal of EdgeVTP: enabling interaction-aware forecasting under realistic E2E measurement. For Table 4, entries marked * use E2E latencies reported by the cited papers (protocols/hardware may differ). All unstarred E2E latencies are measured by us on NVIDIA H100 using the unified protocol in Section \ref{sec:exp_setup}.

\begin{figure}[t]
  \centering
\includegraphics[width=0.75\columnwidth,keepaspectratio]{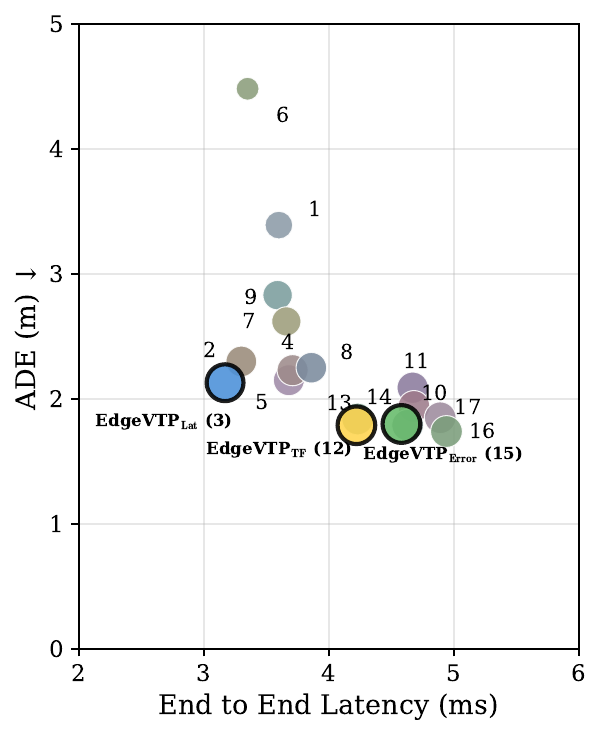}
  \caption{Accuracy--latency trade-off on NGSIM across the EdgeVTP ablation sweep. Each point corresponds to one configuration (extensive ablation results are covered in supplementary material). Highlighted points indicate the three operating points used in the main paper: EdgeVTP$_{\text{Lat}}$ (latency-focused), EdgeVTP$_{\text{TF}}$ (balanced), and EdgeVTP$_{\text{Error}}$ (accuracy-focused). All points report E2E latency under the protocol in Section \ref{sec:exp_setup}}
  \label{fig:ade_e2e_ablation}
\end{figure}

\noindent\textbf{Jetson latency.}
Table~\ref{tab:jetson_latency_combined} reports batch size 1 E2E latency on Jetson Nano (5W/10W) and Jetson Xavier NX (10W 2-core / 20W 6-core). Across devices and power modes, EdgeVTP operates in the tens-of-milliseconds regime, while the VT-Former baseline is substantially slower. The three EdgeVTP operating points offer a clear latency/accuracy spectrum, enabling flexible selection under different power and frame-time budgets.

\begin{table}[h]
\centering
\caption{E2E latency on Jetson devices (batch=1), reported in ms (lower is better). We evaluate Jetson Nano under 10W/5W and Jetson Xavier NX under 10W (2-core) and 20W (6-core) power modes.}
\label{tab:jetson_latency_combined}
\resizebox{\columnwidth}{!}{%
\begin{tabular}{lcc|cc}
\toprule
& \multicolumn{2}{c|}{\textbf{Jetson Nano}} & \multicolumn{2}{c}{\textbf{Jetson Xavier NX}} \\
\cmidrule(lr){2-3}\cmidrule(lr){4-5}
\textbf{Model} & \textbf{10W} & \textbf{5W} & \textbf{10W (2-core)} & \textbf{20W (6-core)} \\
\midrule
VT-Former$_{LH}$ \cite{pazho2024vtformer}            & 1034.26 & 1669.51 & 416.66 & 400.95 \\
STA-LSTM \cite{lin2021stalstm}                   & 51.82   & 102.85  & 29.49  & 30.64  \\
\midrule
\rowcolor[HTML]{D9D9D9}
EdgeVTP$_{\text{Lat}}$   & 27.87 & 48.46 & 14.06 & 11.85 \\
\rowcolor[HTML]{D9D9D9}
EdgeVTP$_{\text{TF}}$    & 38.27 & 66.46 & 19.10 & 16.21 \\
\rowcolor[HTML]{D9D9D9}
EdgeVTP$_{\text{Error}}$ & 37.36 & 65.52 & 19.07 & 16.23 \\
\bottomrule
\end{tabular}%
}
\end{table}


\begin{figure}[b]
\centering

\vspace{-10pt}
\begin{subfigure}[t]{\columnwidth}
  \centering
  \includegraphics[width=0.65\linewidth,trim=0 0 0 3cm,clip]{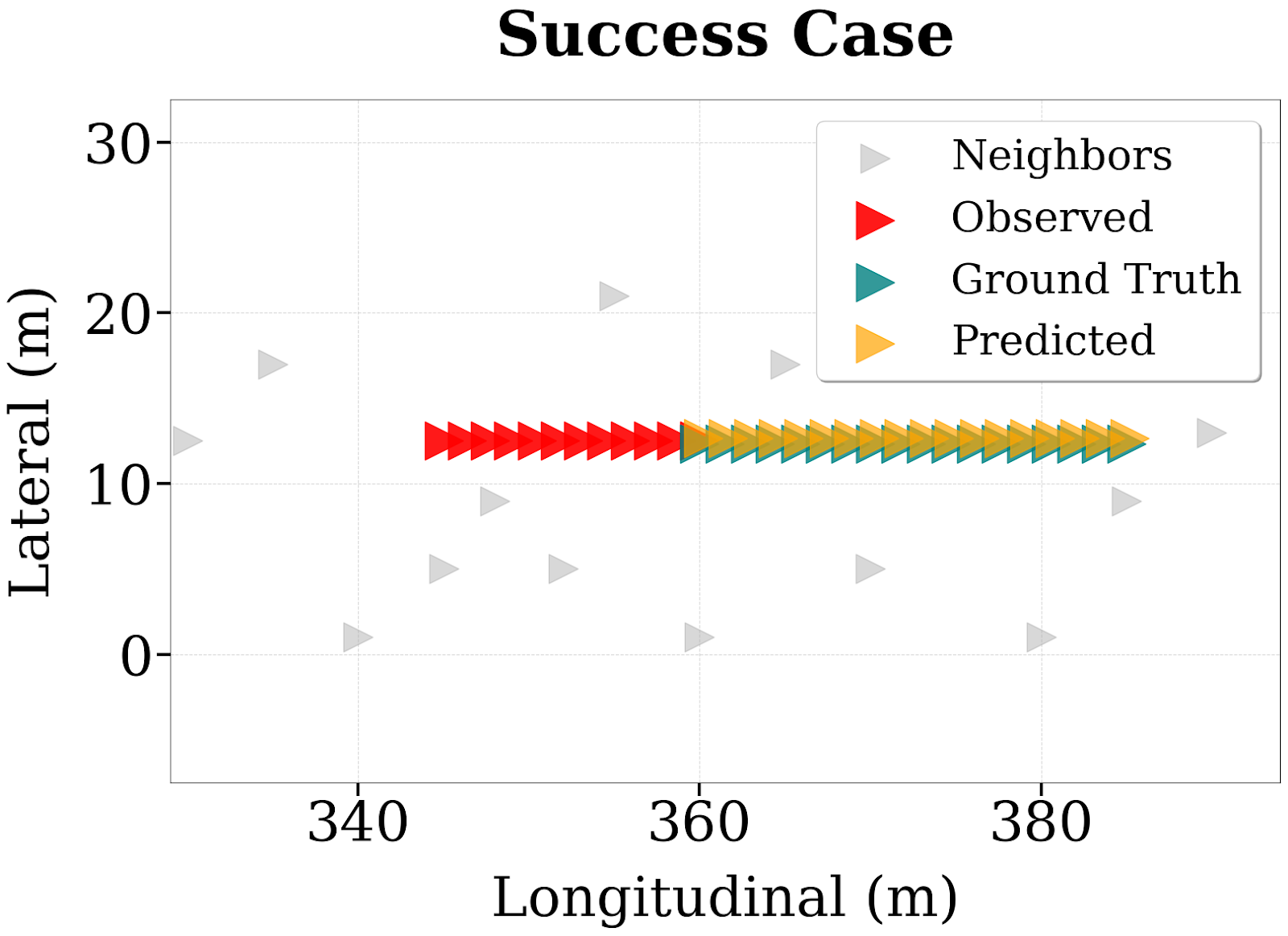}
  \caption{Success Case}
\end{subfigure}\\[-2pt]

\begin{subfigure}[t]{\columnwidth}
  \centering
  \includegraphics[width=0.65\linewidth,trim=0 0 0 3cm,clip]{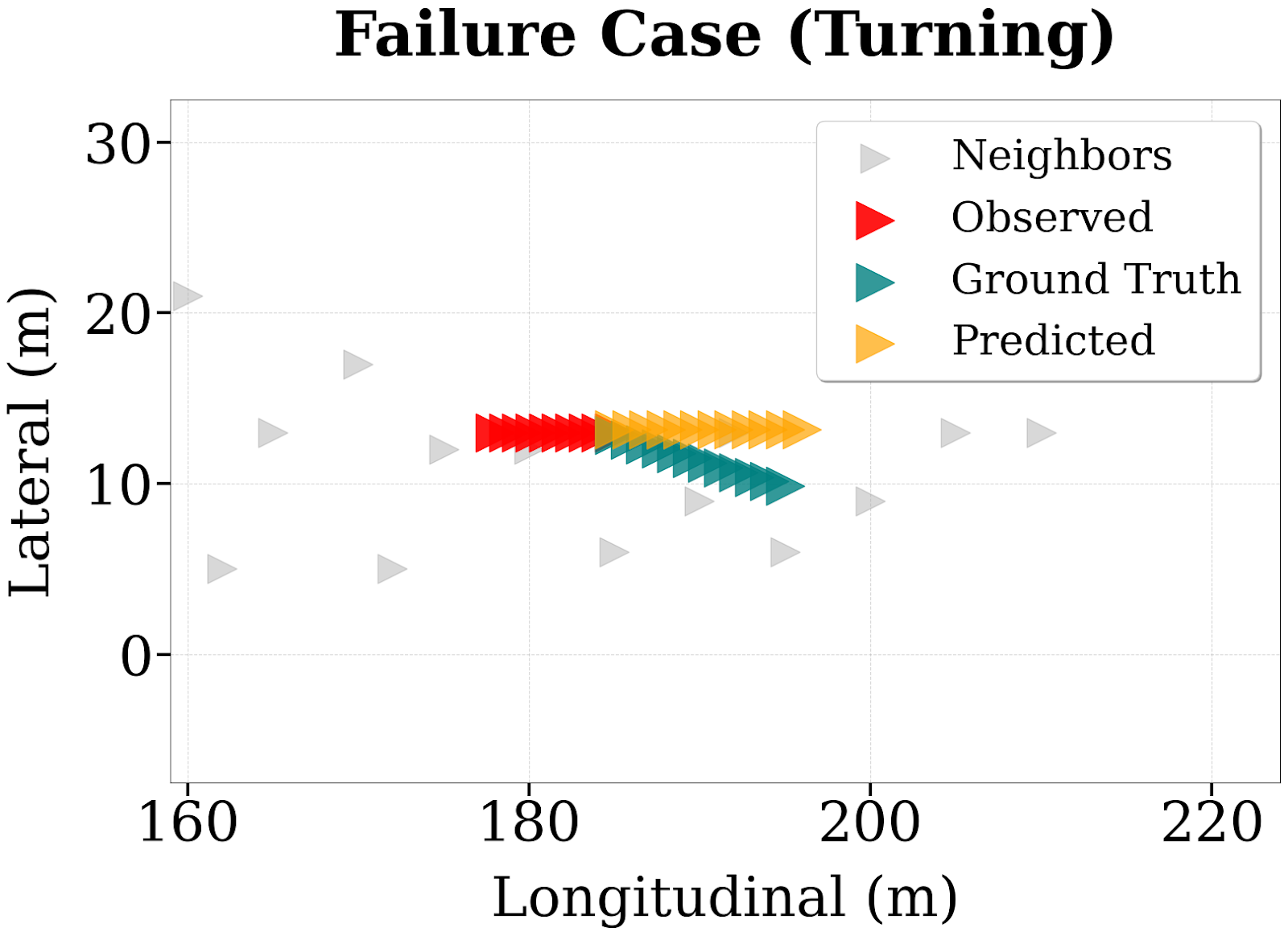}
  \caption{Failure Case (Turning)}
\end{subfigure}\\[-2pt]

\begin{subfigure}[t]{\columnwidth}
  \centering
  \includegraphics[width=0.65\linewidth,trim=0 0 0 3cm,clip]{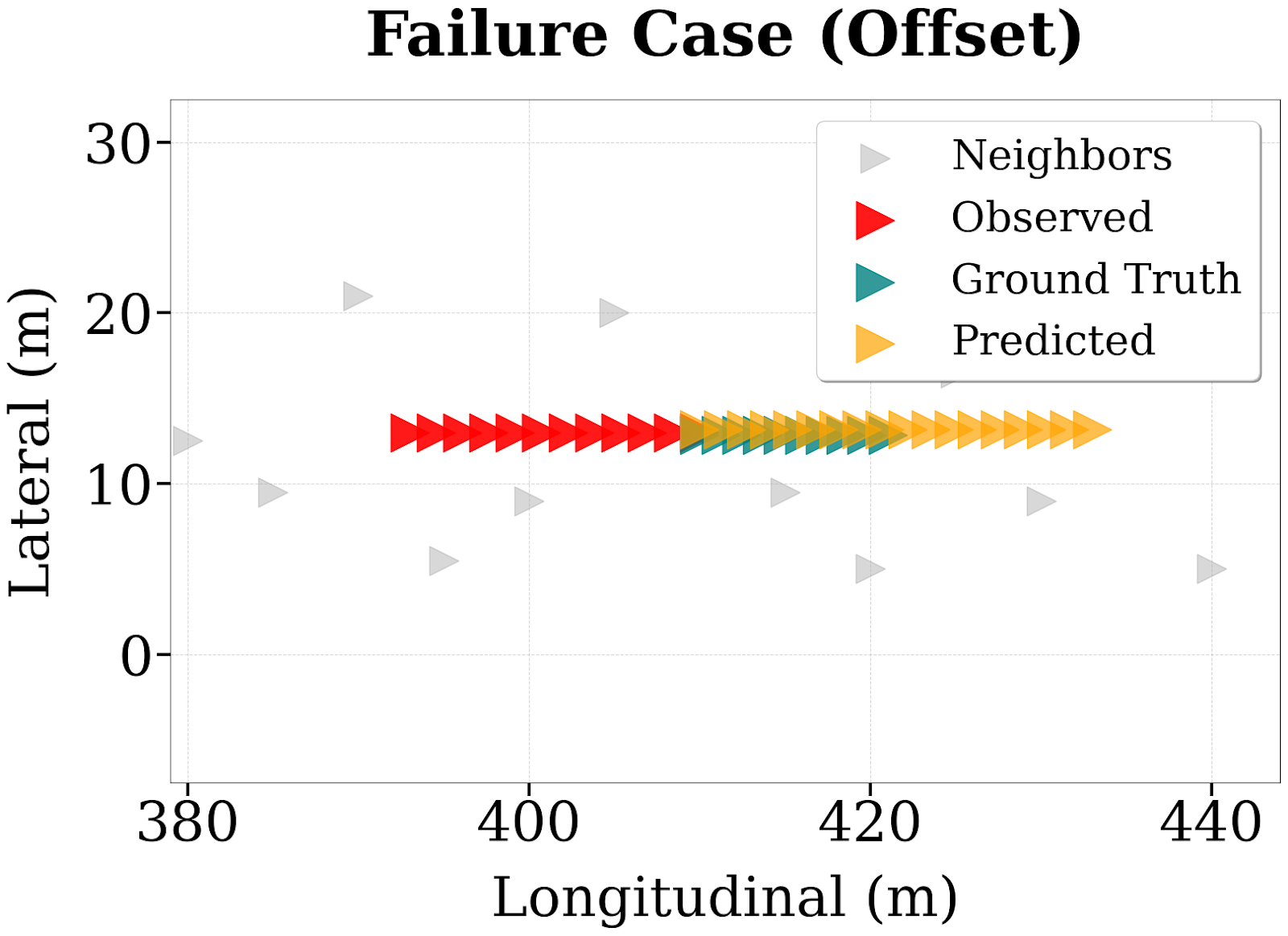}
  \caption{Failure Case (Offset)}
\end{subfigure}

\vspace{-4pt}
\caption{Qualitative results on NGSIM. Observed history is red, predictions are orange, ground truth is teal, and neighbors are gray.}
\label{fig:qualitative}
\vspace{-2pt}
\end{figure}

\noindent\textbf{Qualitative results and failure modes.}
Figure~\ref{fig:qualitative} visualizes typical predictions on NGSIM. We observe successful cases where the model closely matches the ground truth over the full horizon, as well as two common failure modes: (i) lateral offset errors (correct trend but shifted laterally), and (ii) maneuver ambiguity near turns or lane changes, where multiple futures remain plausible from short history alone. These cases motivate future extensions with additional intent cues (e.g., longer context or map topology) while preserving the embedded constraints.

\section{Discussion and Conclusion}
\label{sec:discussion_conclusion}

EdgeVTP addresses fixed-camera highway trajectory prediction under embedded constraints, where end-to-end latency must stay low and stable even as traffic density increases. Our results suggest that for Jetson-class deployment, end-to-end runtime is driven primarily by scene-dependent interaction construction and horizon-scaled decoding overhead, rather than parameter count alone. Bounding the interaction graph with radius gating and a hard top-$K$ cap, together with one-shot B\'ezier decoding, improves the accuracy--latency trade-off under an end-to-end protocol that includes graph construction and trajectory reconstruction. The best operating point varies with viewpoint: on CHD, EdgeVTP achieves the SotA accuracy among the listed methods on both splits, with EdgeVTP$_{\text{TF}}$ best on Eye-level and EdgeVTP$_{\text{Error}}$ best on High-angle \cite{chd}. On NGSIM, EdgeVTP achieves the lowest measured end-to-end latency among the compared methods while remaining competitive in prediction error.


\textbf{Conclusion.} We presented EdgeVTP, an embedded-first predictor for roadside surveillance that combines bounded interaction graphs with one-shot degree-4 B\'ezier decoding. Across CHD, it achieves the best accuracy and lowest latency among the listed methods, and on NGSIM  it achieves the lowest measured end-to-end latency among the compared methods while maintaining competitive accuracy. These results show that making deployment constraints explicit in the architecture can yield practical trajectory prediction for continuous edge-based highway monitoring.

{
    \small
    \bibliographystyle{ieeenat_fullname}
    \bibliography{main}

@String(CVPR= {IEEE Conf. Comput. Vis. Pattern Recog.})

@String(ICCV= {Int. Conf. Comput. Vis.})

@String(ECCV= {Eur. Conf. Comput. Vis.})

@String(BMVC= {Brit. Mach. Vis. Conf.})

@String(ICLR = {Int. Conf. Learn. Represent.})

@String(AAAI = {AAAI})

@String(CVPRW= {IEEE Conf. Comput. Vis. Pattern Recog. Worksh.})

@String(CVPR  = {CVPR})

@String(ICCV  = {ICCV})

@String(ECCV  = {ECCV})

@String(BMVC  =	{BMVC})

@String(ICLR  = {ICLR})

@String(CVPRW= {CVPRW})

@inproceedings{pazho2024vtformer,
  title        = {VT-Former: An Exploratory Study on Vehicle Trajectory Prediction for Highway Surveillance through Graph Isomorphism and Transformer},
  author       = {Pazho, Armin Danesh and Noghre, Ghazal Alinezhad and Katariya, Vinit and Tabkhi, Hamed},
  booktitle    = {Proceedings of the IEEE/CVF Conference on Computer Vision and Pattern Recognition Workshops (CVPRW)},
  year         = {2024}
}

@inproceedings{deo2018csp,
  title        = {Convolutional Social Pooling for Vehicle Trajectory Prediction},
  author       = {Deo, Nachiket and Trivedi, Mohan M.},
  booktitle    = {Proceedings of the IEEE/CVF Conference on Computer Vision and Pattern Recognition Workshops (CVPRW)},
  year         = {2018}
}

@article{deo2018mm,
  title        = {Multi-Modal Trajectory Prediction of Surrounding Vehicles with Maneuver based LSTMs},
  author       = {Deo, Nachiket and Trivedi, Mohan M.},
  journal      = {arXiv preprint arXiv:1805.05499},
  year         = {2018}
}

@inproceedings{salzmann2020trajectronpp,
  title        = {Trajectron++: Dynamically-Feasible Trajectory Forecasting With Heterogeneous Data},
  author       = {Salzmann, Tim and Ivanovic, Boris and Chakravarty, Punarjay and Pavone, Marco},
  booktitle    = {European Conference on Computer Vision (ECCV)},
  year         = {2020}
}

@inproceedings{gajiyang2020vectornet,
  title        = {VectorNet: Encoding HD Maps and Agent Dynamics from Vectorized Representation},
  author       = {Gao, Jiyang and Sun, Chen and Zhao, Hang and Shen, Yi and Anguelov, Dragomir and Li, Congcong and Schmid, Cordelia},
  booktitle    = {Proceedings of the IEEE/CVF Conference on Computer Vision and Pattern Recognition (CVPR)},
  year         = {2020}
}

@inproceedings{liang2020lanegcn,
  title        = {Learning Lane Graph Representations for Motion Forecasting},
  author       = {Liang, Ming and Yang, Bin and Hu, Rui and Chen, Yun and Liao, Renjie and Feng, Song and Urtasun, Raquel},
  booktitle    = {European Conference on Computer Vision (ECCV)},
  year         = {2020}
}

@inproceedings{qi2020nart,
  title        = {Imitative Non-Autoregressive Modeling for Trajectory Forecasting and Imputation},
  author       = {Qi, Mengshi and Qin, Jie and Wu, Yu and Yang, Yi},
  booktitle    = {Proceedings of the IEEE/CVF Conference on Computer Vision and Pattern Recognition (CVPR)},
  year         = {2020}
}

@article{achaji2022pretr,
  title        = {PreTR: Spatio-Temporal Non-Autoregressive Trajectory Prediction Transformer},
  author       = {Achaji, Lina and Barry, Thierno and Fouqueray, Thibault and Moreau, Julien and Aioun, Francois and Charpillet, Francois},
  journal      = {arXiv preprint arXiv:2203.09293},
  year         = {2022}
}

@article{singh2023edgeai,
  title        = {Edge AI: A survey},
  author       = {Singh, Raghubir and Gill, Sukhpal Singh},
  journal      = {Internet of Things and Cyber-Physical Systems},
  volume       = {3},
  pages        = {71--92},
  year         = {2023},
  doi          = {10.1016/j.iotcps.2023.02.004}
}

@article{swaminathan2024jetsonnano,
  title        = {Benchmarking Deep Learning Models on NVIDIA Jetson Nano for Real-Time Systems: An Empirical Investigation},
  author       = {Swaminathan, Tushar Prasanna and Silver, Christopher and Akilan, Thangarajah},
  journal      = {arXiv preprint arXiv:2406.17749},
  year         = {2024}
}

@inproceedings{liao2023bat, 
  title     = {BAT: Behavior-Aware Human-Like Trajectory Prediction for Autonomous Driving},
  author    = {Liao, Haicheng and Li, Zhenning and Shen, Huanming and Zeng, Wenxuan and Liao, Dongping and Li, Guofa and Xu, Chengzhong},
  booktitle = {Proceedings of the AAAI Conference on Artificial Intelligence},
  year      = {2024},
  pages     = {10332--10340},
  doi       = {10.1609/AAAI.V38I9.28900}
}

@article{liao2024hltp,
  title   = {A Cognitive-Based Trajectory Prediction Approach for Autonomous Driving},
  author  = {Liao, Haicheng and Li, Yongkang and Li, Zhenning and Wang, Chengyue and Cui, Zhiyong and Li, Shengbo Eben and Xu, Chengzhong},
  journal = {IEEE Transactions on Intelligent Vehicles},
  year    = {2024},
  doi     = {10.1109/TIV.2024.3376074},
  note    = {Early Access}
}

@article{katariya2022deeptrack, 
  title   = {DeepTrack: Lightweight Deep Learning for Vehicle Trajectory Prediction in Highways},
  author  = {Katariya, Vinit and Baharani, Mohammadreza and Morris, Nichole and Shoghli, Omidreza and Tabkhi, Hamed},
  journal = {IEEE Transactions on Intelligent Transportation Systems},
  year    = {2022},
  volume  = {23},
  number  = {10},
  pages   = {18927--18936},
  doi     = {10.1109/TITS.2022.3172015}
}

@inproceedings{li2019grip, 
  title     = {GRIP: Graph-based Interaction-aware Trajectory Prediction},
  author    = {Li, Xin and Ying, Xiaowen and Chuah, Mooi Choo},
  booktitle = {2019 IEEE Intelligent Transportation Systems Conference (ITSC)},
  year      = {2019},
  pages     = {3960--3966},
  doi       = {10.1109/ITSC.2019.8917228}
}

@inproceedings{mohamed2020socialstgcnn,
  title     = {Social-STGCNN: A Social Spatio-Temporal Graph Convolutional Neural Network for Human Trajectory Prediction},
  author    = {Mohamed, Abduallah A. and Qian, Kun and Elhoseiny, Mohamed and Claudel, Christian G.},
  booktitle = {Proceedings of the IEEE/CVF Conference on Computer Vision and Pattern Recognition (CVPR)},
  year      = {2020},
  pages     = {14412--14420},
  doi       = {10.1109/CVPR42600.2020.01443}
}

@article{lin2021stalstm,
  title   = {Vehicle Trajectory Prediction Using LSTMs with Spatial-Temporal Attention Mechanisms},
  author  = {Lin, Lei and Li, Weizi and Bi, Huikun and Qin, Lingqiao},
  journal = {IEEE Intelligent Transportation Systems Magazine},
  year    = {2021},
  doi     = {10.1109/MITS.2021.3049404}
}

@inproceedings{velickovic2018gat,
  title     = {Graph Attention Networks},
  author    = {Veli{\v{c}}kovi{\'c}, Petar and Cucurull, Guillem and Casanova, Arantxa and Romero, Adriana and Li{\`o}, Pietro and Bengio, Yoshua},
  booktitle = {International Conference on Learning Representations (ICLR)},
  year      = {2018},
  note      = {Poster}
}

@article{gou2021kd,
  title   = {Knowledge Distillation: A Survey},
  author  = {Gou, Jianping and Yu, Baosheng and Maybank, Stephen J. and Tao, Dacheng},
  journal = {International Journal of Computer Vision},
  year    = {2021},
  volume  = {129},
  number  = {6},
  pages   = {1789--1819},
  doi     = {10.1007/s11263-021-01453-z}
}

@inproceedings{hug2020probbezier,
  title     = {Introducing Probabilistic B{\'e}zier Curves for {N}-Step Sequence Prediction},
  author    = {Hug, Ronny and H{\"u}bner, Wolfgang and Arens, Michael},
  booktitle = {Proceedings of the AAAI Conference on Artificial Intelligence},
  year      = {2020},
  volume    = {34},
  number    = {06},
  pages     = {10162--10169},
  url       = {https://ojs.aaai.org/index.php/AAAI/article/view/6576}
}

@inproceedings{prutsch2024efficient,
  title     = {Efficient Motion Prediction: A Lightweight \& Accurate Trajectory Prediction Model With Fast Training and Inference Speed},
  author    = {Prutsch, Alexander and Bischof, Horst and Possegger, Horst},
  booktitle = {2024 IEEE/RSJ International Conference on Intelligent Robots and Systems (IROS)},
  pages     = {9411--9417},
  year      = {2024},
  doi       = {10.1109/IROS58592.2024.10802425}
}

@article{chd,
  title={A pov-based highway vehicle trajectory dataset and prediction architecture},
  author={Katariya, Vinit and Noghre, Ghazal Alinezhad and Pazho, Armin Danesh and Tabkhi, Hamed},
  journal={IEEE Transactions on Intelligent Transportation Systems},
  volume={25},
  number={10},
  pages={13136--13146},
  year={2024},
  publisher={IEEE}
}

@article{li2019grip++,
  title={Grip++: Enhanced graph-based interaction-aware trajectory prediction for autonomous driving},
  author={Li, Xin and Ying, Xiaowen and Chuah, Mooi Choo},
  journal={arXiv preprint arXiv:1907.07792},
  year={2019}
}

@inproceedings{alinezhad2023pishgu, 
  title={Pishgu: Universal Path Prediction Network Architecture for Real-time Cyber-physical Edge Systems},
  author={Alinezhad Noghre, Ghazal and Katariya, Vinit and Danesh Pazho, Armin and Neff, Christopher and Tabkhi, Hamed},
  booktitle={Proceedings of the ACM/IEEE 14th International Conference on Cyber-Physical Systems (with CPS-IoT Week 2023)},
  pages={88--97},
  year={2023}
}

@InProceedings{Shi_2023_ICCV,
  author    = {Shi, Liushuai and Wang, Le and Zhou, Sanping and Hua, Gang},
  title     = {Trajectory Unified Transformer for Pedestrian Trajectory Prediction},
  booktitle = {Proceedings of the IEEE/CVF International Conference on Computer Vision (ICCV)},
  month     = {October},
  year      = {2023},
  pages     = {9675-9684}
}

@InProceedings{Belhadi_2025_ICCV,
  author    = {Belhadi, Asma and Djenouri, Youcef and Belbachir, Ahmed Nabil},
  title     = {LightPrune: Latency-Aware Structured Pruning for Efficient Deep Inference on Embedded Devices},
  booktitle = {Proceedings of the IEEE/CVF International Conference on Computer Vision (ICCV) Workshops},
  month     = {October},
  year      = {2025},
  pages     = {1688-1697}
}

@InProceedings{Chen_2024_CVPR,
  author    = {Chen, Tse-Wei and Tao, Wei and Zhao, Dongyue and Mima, Kazuhiro and Ito, Tadayuki and Osa, Kinya and Kato, Masami},
  title     = {Dedicated Inference Engine and Binary-Weight Neural Networks for Lightweight Instance Segmentation},
  booktitle = {Proceedings of the IEEE/CVF Conference on Computer Vision and Pattern Recognition (CVPR) Workshops},
  month     = {June},
  year      = {2024},
  pages     = {2101-2110}
}

@InProceedings{Prabhune_2024_CVPR,
  author    = {Prabhune, Omkar and Chen, Tianen and Kim, Younghyun},
  title     = {Content-aware Input Scaling and Deep Learning Computation Offloading for Low-Latency Embedded Vision},
  booktitle = {Proceedings of the IEEE/CVF Conference on Computer Vision and Pattern Recognition (CVPR) Workshops},
  month     = {June},
  year      = {2024},
  pages     = {2218-2226}
}

@electronic{NGSIM_US101,
  author        = "James Colyar and John Halkias",
  title         = " Next Generation SIMulation ({NGSIM}), {US} {H}ighway-101 dataset. {FHWA-HRT}-07-030.",
  url           = "https://www.fhwa.dot.gov/publications/research/operations/07030/",
  year          = "2007"
}

@electronic{NGSIM_i80, 
  author        = "James Colyar and John Halkias",
  title         = "Next Generation SIMulation ({NGSIM}), {I}nterstate 80 Freeway Dataset. {FHWA-HRT}-06-137",
  url           = "https://www.fhwa.dot.gov/publications/research/operations/06137/",
  year          = "2006"
}

@article{chen2022vehicle,
  title={Vehicle trajectory prediction based on intention-aware non-autoregressive transformer with multi-attention learning for Internet of Vehicles},
  author={Chen, Xiaobo and Zhang, Huanjia and Zhao, Feng and Cai, Yingfeng and Wang, Hai and Ye, Qiaolin},
  journal={IEEE Transactions on Instrumentation and Measurement},
  volume={71},
  pages={1--12},
  year={2022},
  publisher={IEEE}
}

@article{gao2023dual,
  title={Dual transformer based prediction for lane change intentions and trajectories in mixed traffic environment},
  author={Gao, Kai and Li, Xunhao and Chen, Bin and Hu, Lin and Liu, Jian and Du, Ronghua and Li, Yongfu},
  journal={IEEE Transactions on Intelligent Transportation Systems},
  volume={24},
  number={6},
  pages={6203--6216},
  year={2023},
  publisher={IEEE}
}

@article{vaswani2017attention,
  title={Attention is all you need},
  author={Vaswani, Ashish and Shazeer, Noam and Parmar, Niki and Uszkoreit, Jakob and Jones, Llion and Gomez, Aidan N and Kaiser, {\L}ukasz and Polosukhin, Illia},
  journal={Advances in neural information processing systems},
  volume={30},
  year={2017}
}

@article{hinton2015distilling,
  title={Distilling the knowledge in a neural network},
  author={Hinton, Geoffrey and Vinyals, Oriol and Dean, Jeff},
  journal={arXiv preprint arXiv:1503.02531},
  year={2015}
}

@inproceedings{alahi2016social,
  title={Social lstm: Human trajectory prediction in crowded spaces},
  author={Alahi, Alexandre and Goel, Kratarth and Ramanathan, Vignesh and Robicquet, Alexandre and Fei-Fei, Li and Savarese, Silvio},
  booktitle={Proceedings of the IEEE conference on computer vision and pattern recognition},
  pages={961--971},
  year={2016}
}

@inproceedings{gupta2018social,
  title={Social gan: Socially acceptable trajectories with generative adversarial networks},
  author={Gupta, Agrim and Johnson, Justin and Fei-Fei, Li and Savarese, Silvio and Alahi, Alexandre},
  booktitle={Proceedings of the IEEE conference on computer vision and pattern recognition},
  pages={2255--2264},
  year={2018}
}

@article{ngiam2021scene,
  title={Scene transformer: A unified architecture for predicting multiple agent trajectories},
  author={Ngiam, Jiquan and Caine, Benjamin and Vasudevan, Vijay and Zhang, Zhengdong and Chiang, Hao-Tien Lewis and Ling, Jeffrey and Roelofs, Rebecca and Bewley, Alex and Liu, Chenxi and Venugopal, Ashish and others},
  journal={arXiv preprint arXiv:2106.08417},
  year={2021}
}

@inproceedings{yuan2021agentformer,
  title={Agentformer: Agent-aware transformers for socio-temporal multi-agent forecasting},
  author={Yuan, Ye and Weng, Xinshuo and Ou, Yanglan and Kitani, Kris M},
  booktitle={Proceedings of the IEEE/CVF international conference on computer vision},
  pages={9813--9823},
  year={2021}
}

@inproceedings{zhou2022hivt,
  title={Hivt: Hierarchical vector transformer for multi-agent motion prediction},
  author={Zhou, Zikang and Ye, Luyao and Wang, Jianping and Wu, Kui and Lu, Kejie},
  booktitle={Proceedings of the IEEE/CVF conference on computer vision and pattern recognition},
  pages={8823--8833},
  year={2022}
}

@inproceedings{huang2025trajectory,
  title={Trajectory mamba: Efficient attention-mamba forecasting model based on selective ssm},
  author={Huang, Yizhou and Cheng, Yihua and Wang, Kezhi},
  booktitle={Proceedings of the Computer Vision and Pattern Recognition Conference},
  pages={12058--12067},
  year={2025}
}

@inproceedings{caesar2020nuscenes,
  title={nuscenes: A multimodal dataset for autonomous driving},
  author={Caesar, Holger and Bankiti, Varun and Lang, Alex H and Vora, Sourabh and Liong, Venice Erin and Xu, Qiang and Krishnan, Anush and Pan, Yu and Baldan, Giancarlo and Beijbom, Oscar},
  booktitle={Proceedings of the IEEE/CVF conference on computer vision and pattern recognition},
  pages={11621--11631},
  year={2020}
}

@inproceedings{sun2020scalability,
  title={Scalability in perception for autonomous driving: Waymo open dataset},
  author={Sun, Pei and Kretzschmar, Henrik and Dotiwalla, Xerxes and Chouard, Aurelien and Patnaik, Vijaysai and Tsui, Paul and Guo, James and Zhou, Yin and Chai, Yuning and Caine, Benjamin and others},
  booktitle={Proceedings of the IEEE/CVF conference on computer vision and pattern recognition},
  pages={2446--2454},
  year={2020}
}

@article{wilson2023argoverse,
  title={Argoverse 2: Next generation datasets for self-driving perception and forecasting},
  author={Wilson, Benjamin and Qi, William and Agarwal, Tanmay and Lambert, John and Singh, Jagjeet and Khandelwal, Siddhesh and Pan, Bowen and Kumar, Ratnesh and Hartnett, Andrew and Pontes, Jhony Kaesemodel and others},
  journal={arXiv preprint arXiv:2301.00493},
  year={2023}
}

@inproceedings{chang2019argoverse,
  title={Argoverse: 3d tracking and forecasting with rich maps},
  author={Chang, Ming-Fang and Lambert, John and Sangkloy, Patsorn and Singh, Jagjeet and Bak, Slawomir and Hartnett, Andrew and Wang, De and Carr, Peter and Lucey, Simon and Ramanan, Deva and others},
  booktitle={Proceedings of the IEEE/CVF conference on computer vision and pattern recognition},
  pages={8748--8757},
  year={2019}
}

@inproceedings{alibeigi2023zenseact,
  title={Zenseact open dataset: A large-scale and diverse multimodal dataset for autonomous driving},
  author={Alibeigi, Mina and Ljungbergh, William and Tonderski, Adam and Hess, Georg and Lilja, Adam and Lindstr{\"o}m, Carl and Motorniuk, Daria and Fu, Junsheng and Widahl, Jenny and Petersson, Christoffer},
  booktitle={Proceedings of the IEEE/CVF International Conference on Computer Vision},
  pages={20178--20188},
  year={2023}
}

@inproceedings{giannakeris2018speed,
  title={Speed estimation and abnormality detection from surveillance cameras},
  author={Giannakeris, Panagiotis and Kaltsa, Vagia and Avgerinakis, Konstantinos and Briassouli, Alexia and Vrochidis, Stefanos and Kompatsiaris, Ioannis},
  booktitle={Proceedings of the IEEE Conference on Computer Vision and Pattern Recognition Workshops},
  pages={93--99},
  year={2018}
}

@inproceedings{wei2018unsupervised,
  title={Unsupervised anomaly detection for traffic surveillance based on background modeling},
  author={Wei, JiaYi and Zhao, JianFei and Zhao, YanYun and Zhao, ZhiCheng},
  booktitle={Proceedings of the IEEE conference on computer vision and pattern recognition workshops},
  pages={129--136},
  year={2018}
}

@inproceedings{verma2024etram,
  title={etram: Event-based traffic monitoring dataset},
  author={Verma, Aayush Atul and Chakravarthi, Bharatesh and Vaghela, Arpitsinh and Wei, Hua and Yang, Yezhou},
  booktitle={Proceedings of the IEEE/CVF conference on computer vision and pattern recognition},
  pages={22637--22646},
  year={2024}
}

@article{hasanujjaman2023sensor, 
  title={Sensor fusion in autonomous vehicle with traffic surveillance camera system: detection, localization, and AI networking},
  author={Hasanujjaman, Muhammad and Chowdhury, Mostafa Zaman and Jang, Yeong Min},
  journal={Sensors},
  volume={23},
  number={6},
  pages={3335},
  year={2023},
  publisher={MDPI}
}

@article{qiu2024intelligent,
  title={Intelligent highway adaptive lane learning system in multiple rois of surveillance camera video},
  author={Qiu, Mei and Christopher, Lauren and Chien, Stanley Yung-Ping and Chen, Yaobin},
  journal={IEEE Transactions on Intelligent Transportation Systems},
  volume={25},
  number={8},
  pages={8591--8601},
  year={2024},
  publisher={IEEE}
}

@inproceedings{xu2024adapting, 
  title={Adapting to length shift: Flexilength network for trajectory prediction},
  author={Xu, Yi and Fu, Yun},
  booktitle={Proceedings of the IEEE/CVF Conference on Computer Vision and Pattern Recognition},
  pages={15226--15237},
  year={2024}
}

@inproceedings{zhang2024oostraj,
  title={Oostraj: Out-of-sight trajectory prediction with vision-positioning denoising},
  author={Zhang, Haichao and Xu, Yi and Lu, Hongsheng and Shimizu, Takayuki and Fu, Yun},
  booktitle={Proceedings of the IEEE/CVF Conference on Computer Vision and Pattern Recognition},
  pages={14802--14811},
  year={2024}
}

@inproceedings{tang2024hpnet, 
  title={Hpnet: Dynamic trajectory forecasting with historical prediction attention},
  author={Tang, Xiaolong and Kan, Meina and Shan, Shiguang and Ji, Zhilong and Bai, Jinfeng and Chen, Xilin},
  booktitle={Proceedings of the IEEE/CVF conference on computer vision and pattern recognition},
  pages={15261--15270},
  year={2024}
}

@INPROCEEDINGS {10657975,
author = { Bae, Inhwan and Lee, Junoh and Jeon, Hae-Gon },
booktitle = { 2024 IEEE/CVF Conference on Computer Vision and Pattern Recognition (CVPR) },
title = {{ Can Language Beat Numerical Regression? Language-Based Multimodal Trajectory Prediction }},
year = {2024},
volume = {},
ISSN = {},
pages = {753-766},
doi = {10.1109/CVPR52733.2024.00078},
url = {https://doi.ieeecomputersociety.org/10.1109/CVPR52733.2024.00078},
publisher = {IEEE Computer Society},
address = {Los Alamitos, CA, USA},
month =Jun}

@inproceedings{pourkeshavarz2024cadet,
  title={Cadet: a causal disentanglement approach for robust trajectory prediction in autonomous driving},
  author={Pourkeshavarz, Mozhgan and Zhang, Junrui and Rasouli, Amir},
  booktitle={Proceedings of the IEEE/CVF Conference on Computer Vision and Pattern Recognition},
  pages={14874--14884},
  year={2024}
}

@inproceedings{krajewski2018highd,
  title={The highd dataset: A drone dataset of naturalistic vehicle trajectories on german highways for validation of highly automated driving systems},
  author={Krajewski, Robert and Bock, Julian and Kloeker, Laurent and Eckstein, Lutz},
  booktitle={2018 21st international conference on intelligent transportation systems (ITSC)},
  pages={2118--2125},
  year={2018},
  organization={IEEE}
}

@inproceedings{moers2022exid,
  title={The exid dataset: A real-world trajectory dataset of highly interactive highway scenarios in germany},
  author={Moers, Tobias and Vater, Lennart and Krajewski, Robert and Bock, Julian and Zlocki, Adrian and Eckstein, Lutz},
  booktitle={2022 IEEE Intelligent Vehicles Symposium (IV)},
  pages={958--964},
  year={2022},
  organization={IEEE}
}

@article{li2014stop, 
  title={Stop-and-go traffic analysis: Theoretical properties, environmental impacts and oscillation mitigation},
  author={Li, Xiaopeng and Cui, Jianxun and An, Shi and Parsafard, Mohsen},
  journal={Transportation Research Part B: Methodological},
  volume={70},
  pages={319--339},
  year={2014},
  publisher={Elsevier}
}

@article{grimm2025goal, 
  title={Goal-based Trajectory Prediction for improved Cross-Dataset Generalization},
  author={Grimm, Daniel and Abouelazm, Ahmed and Z{\"o}llner, J Marius},
  journal={arXiv preprint arXiv:2507.18196},
  year={2025}
}

@InProceedings{Ozer_2014_CVPR_Workshops,
  author    = {Ozer, Burak and Wolf, Marilyn},
  title     = {A Train Station Surveillance System: Challenges and Solutions},
  booktitle = {Proceedings of the IEEE Conference on Computer Vision and Pattern Recognition (CVPR) Workshops},
  month     = {June},
  year      = {2014}
}

@article{Apewokin_RealTimeAdaptiveBackground_2011,
  title   = {Real-Time Adaptive Background Modeling for Multicore Embedded Systems},
  author  = {Apewokin, Senyo and Valentine, Brian and Choi, Jee and Wills, Linda and Wills, Scott},
  journal = {Journal of Signal Processing Systems},
  year    = {2011},
  volume  = {62},
  pages   = {65--76},
  doi     = {10.1007/s11265-008-0298-z}
}

@InProceedings{Chandorkar_2025_ICCV,
  author    = {Chandorkar, Adwait and Tercan, Hasan and Meisen, Tobias},
  title     = {Rethinking Backbone Design for Lightweight 3D Object Detection in LiDAR},
  booktitle = {Proceedings of the IEEE/CVF International Conference on Computer Vision (ICCV) Workshops},
  month     = {October},
  year      = {2025},
  pages     = {1698-1706}
}

@InProceedings{Cigla_2018_CVPR_Workshops,
  author    = {Cigla, Cevahir and Thakker, Rohan and Matthies, Larry},
  title     = {Onboard Stereo Vision for Drone Pursuit or Sense and Avoid},
  booktitle = {Proceedings of the IEEE Conference on Computer Vision and Pattern Recognition (CVPR) Workshops},
  month     = {June},
  year      = {2018}
}

@inproceedings{tang2019cityflow,
  title     = {CityFlow: A City-Scale Benchmark for Multi-Target Multi-Camera Vehicle Tracking and Re-Identification},
  author    = {Tang, Zheng and Naphade, Milind and Liu, Ming-Yu and Yang, Xiaodong and Birchfield, Stan and Wang, Shuo and Kumar, Ratnesh and Anastasiu, David and Hwang, Jenq-Neng},
  booktitle = {Proceedings of the IEEE/CVF Conference on Computer Vision and Pattern Recognition (CVPR)},
  year      = {2019},
  doi       = {10.1109/CVPR.2019.00900}
}

@inproceedings{naphade2019aicity,
  title     = {The 2019 AI City Challenge},
  author    = {Naphade, Milind and Tang, Zheng and Chang, Ming-Ching and others},
  booktitle = {Proceedings of the IEEE/CVF Conference on Computer Vision and Pattern Recognition Workshops (CVPRW)},
  year      = {2019}
}

@inproceedings{naphade2020aicity,
  title     = {The 4th AI City Challenge},
  author    = {Naphade, Milind and Wang, Shuo and Anastasiu, David and Tang, Zheng and Chang, Ming-Ching and Yang, Xiaodong and Zheng, Liang and Sharma, Anuj and Chellappa, Rama and Chakraborty, Pranamesh},
  booktitle = {Proceedings of the IEEE/CVF Conference on Computer Vision and Pattern Recognition Workshops (CVPRW)},
  year      = {2020},
  eprint    = {2004.14619},
  archivePrefix = {arXiv}
}

@article{wen2020uadetrac,
  title   = {UA-DETRAC: A New Benchmark and Protocol for Multi-Object Detection and Tracking},
  author  = {Wen, Longyin and Du, Dawei and Cai, Zhaowei and Lei, Zhen and Chang, Ming-Ching and Qi, Honggang and Lim, Jongwoo and Yang, Ming-Hsuan and Lyu, Siwei},
  journal = {Computer Vision and Image Understanding},
  volume  = {193},
  pages   = {102907},
  year    = {2020},
  doi     = {10.1016/j.cviu.2020.102907}
}

@inproceedings{dubska2014trafficcalib,
  title     = {Automatic Camera Calibration for Traffic Understanding},
  author    = {Dubsk{\'a}, Mark{\'e}ta and Sochor, Jakub and Herout, Adam},
  booktitle = {Proceedings of the British Machine Vision Conference (BMVC)},
  year      = {2014}
}

@article{sochor2017trafficcalib,
  title   = {Traffic Surveillance Camera Calibration by 3D Model Bounding Box Alignment for Accurate Vehicle Speed Measurement},
  author  = {Sochor, Jakub and Jur{\'a}nek, Roman and Herout, Adam},
  journal = {arXiv preprint arXiv:1702.06451},
  year    = {2017}
}

@inproceedings{lee2017desire,
  title     = {DESIRE: Distant Future Prediction in Dynamic Scenes with Interacting Agents},
  author    = {Lee, Namhoon and Choi, Wongun and Vernaza, Paul and Choy, Christopher B. and Torr, Philip H. S. and Chandraker, Manmohan},
  booktitle = {Proceedings of the IEEE Conference on Computer Vision and Pattern Recognition (CVPR)},
  year      = {2017}
}

@inproceedings{phanminh2020covernet,
  title={Covernet: Multimodal behavior prediction using trajectory sets},
  author={Phan-Minh, Tung and Grigore, Elena Corina and Boulton, Freddy A and Beijbom, Oscar and Wolff, Eric M},
  booktitle={Proceedings of the IEEE/CVF conference on computer vision and pattern recognition},
  pages={14074--14083},
  year={2020}
}

@article{chai2020multipath,
  title   = {MultiPath: Multiple Probabilistic Anchor Trajectory Hypotheses for Behavior Prediction},
  author  = {Chai, Yuning and Sapp, Benjamin and Bansal, Mayank and Anguelov, Dragomir},
  journal = {arXiv preprint arXiv:1910.05449},
  year    = {2020}
}

@article{zhao2020tnt,
  title   = {TNT: Target-driven Trajectory Prediction},
  author  = {Zhao, Hang and others},
  journal = {arXiv preprint arXiv:2008.08294},
  year    = {2020}
}

@inproceedings{gu2021densetnt,
  title     = {DenseTNT: End-to-End Trajectory Prediction From Dense Goal Sets},
  author    = {Gu, Junru and Sun, Chen and Zhao, Hang},
  booktitle = {Proceedings of the IEEE/CVF International Conference on Computer Vision (ICCV)},
  year      = {2021},
  doi       = {10.1109/ICCV48922.2021.01502}
}

@article{shi2022mtr,
  title   = {Motion Transformer with Global Intention Localization and Local Movement Refinement},
  author  = {Shi, Shaoshuai and Jiang, Li and Dai, Dengxin and Schiele, Bernt},
  journal = {arXiv preprint arXiv:2209.13508},
  year    = {2022}
}

@article{nayakanti2022wayformer,
  title   = {Wayformer: Motion Forecasting via Simple \& Efficient Attention Networks},
  author  = {Nayakanti, Nigamaa and Al-Rfou, Rami and Zhou, Aurick and Goel, Kratarth and Refaat, Khaled S. and Sapp, Benjamin},
  journal = {arXiv preprint arXiv:2207.05844},
  year    = {2022}
}

@inproceedings{zhang2023hptr,
  title     = {Real-Time Motion Prediction via Heterogeneous Polyline Transformer with Relative Pose Encoding},
  author    = {Zhang, Zhejun and Liniger, Alexander and Sakaridis, Christos and Yu, Fisher and Van Gool, Luc},
  booktitle = {Advances in Neural Information Processing Systems (NeurIPS)},
  year      = {2023},
  eprint    = {2310.12970},
  archivePrefix = {arXiv}
}

@inproceedings{lafage2025hltens,
  title     = {Hierarchical Light Transformer Ensembles for Multimodal Trajectory Forecasting},
  author    = {Lafage, Adrien and Barbier, Mathieu and Franchi, Gianni and Filliat, David},
  booktitle = {IEEE/CVF Winter Conference on Applications of Computer Vision (WACV)},
  year      = {2025},
  doi       = {10.1109/WACV61041.2025.00171},
  eprint    = {2403.17678},
  archivePrefix = {arXiv}
}

@inproceedings{reddi2020mlperf,
  title     = {MLPerf Inference Benchmark},
  author    = {Reddi, Vijay Janapa and others},
  booktitle = {Proceedings of the ACM/IEEE 47th Annual International Symposium on Computer Architecture (ISCA)},
  year      = {2020},
  doi       = {10.1109/ISCA45697.2020.00045},
  eprint    = {1911.02549},
  archivePrefix = {arXiv}
}

@inproceedings{chen2018tvm,
  title     = {TVM: An Automated End-to-End Optimizing Compiler for Deep Learning},
  author    = {Chen, Tianqi and Moreau, Thierry and others},
  booktitle = {USENIX Symposium on Operating Systems Design and Implementation (OSDI)},
  year      = {2018},
  eprint    = {1802.04799},
  archivePrefix = {arXiv}
}

@misc{nvidia_tensorrt,
  title        = {NVIDIA TensorRT Documentation},
  howpublished = {\url{https://docs.nvidia.com/deeplearning/tensorrt/latest/index.html}},
  note         = {Accessed 2026-03-04}
}

@article{jeong2022jedi,
  title   = {TensorRT-Based Framework and Optimization Methodology for Deep Learning Inference on Jetson Boards},
  author  = {Jeong, E. and Kim, J. and Ha, S.},
  journal = {ACM Transactions on Embedded Computing Systems},
  year    = {2022},
  doi     = {10.1145/3508391}
}
}

\clearpage
\FloatBarrier
\appendix


\section{Ablation Study and Operating-Point Selection}
\label{sec:supp_ablation}

We ablate EdgeVTP along three design knobs that directly control the accuracy-latency behavior on NGSIM: (i) the interaction radius $r$ (meters), which determines the spatial extent of neighbor interactions; (ii) a hard top-$K$ neighbor cap, which bounds the number of edges per agent and yields predictable message-passing cost; and (iii) whether residual separation is enabled (Residual = Y/N), which increases capacity with an additional compute and memory overhead. Unless stated otherwise, all settings share the same training protocol, Transformer decoder configuration, and one-shot B\'ezier head, and we report both model-only and end-to-end (E2E) latency, where E2E includes graph construction and trajectory reconstruction.

Table~\ref{tab:supp_ablation_full} reports the complete sweep, indexed by serial ID. As expected, increasing $K$ and $r$ tends to increase interaction cost by admitting more neighbors, while enabling residual separation improves error rate at the cost of higher model-only and E2E runtime. Rather than selecting a single configuration based purely on ADE/FDE, we identify three representative operating points that span the practical accuracy-latency spectrum and use them consistently throughout the paper (including Results Section in main paper).

From the sweep, we select three operating points: a latency-focused setting (EdgeVTP$_{\text{Lat}}$), an accuracy-focused setting (EdgeVTP$_{\text{Error}}$), and a balanced knee-point setting (EdgeVTP$_{\text{TF}}$). We refer to the knee point as the configuration that achieves a strong accuracy gain over the latency-focused setting with only a modest additional latency, while avoiding the diminishing returns of more expensive configurations. Table~\ref{tab:supp_ablation_selected} summarizes these selections and Table~\ref{tab:supp_ablation_full} highlights them in color (blue/yellow/green).

\noindent\textbf{Trend summary.}
Across both residual and non-residual variants, $K$ primarily controls latency by bounding the per-agent neighbor count, while $r$ controls the interaction \emph{scope} and can improve error rate when longer-range interactions are informative. Residual separation consistently improves error but increases model-only latency, which in turn raises E2E latency.

\begin{table}[b]
\centering
\caption{Selected operating points from the ablation sweep (NGSIM).}
\label{tab:supp_ablation_selected}
\resizebox{\columnwidth}{!}{%
\begin{tabular}{@{}lcccccc@{}}
\toprule
\textbf{Choice} & \textbf{ID} & \textbf{$r$} & \textbf{$K$} & \textbf{Residual} & \textbf{ADE} & \textbf{E2E (ms)} \\
\midrule
\rowcolor{latblue}
Best latency   & 3  & 20 & 16 & N & 2.13 & 3.17 \\
\rowcolor{tradeyellow}
Best trade-off & 12 & 20 & 16 & Y & 1.89 & 4.30 \\
\rowcolor{errgreen}
Best Error       & 15 & 30 & 16 & Y & 1.85 & 4.58 \\
\bottomrule
\end{tabular}%
}
\end{table}

\begin{table*}[t]
\centering
\caption{Ablation sweep on NGSIM (indexed by serial ID). Errors are in meters; latency is in milliseconds. Blue: best latency. Yellow: best trade-off. Green: best ADE.}
\label{tab:supp_ablation_full}
\resizebox{0.8\textwidth}{!}{%
\begin{tabular}{c c c c|c c|c c c c c|c c c c}
\toprule
\textbf{ID} &
\textbf{$r$} &
\textbf{$K$} &
\shortstack{\textbf{Residual}} &
\textbf{ADE} & \textbf{FDE} &
\multicolumn{5}{c|}{\textbf{RMSE (m)}} &
\shortstack{\textbf{AVG}} &
\shortstack{\textbf{Params}} &
\shortstack{\textbf{E2E}} &
\shortstack{\textbf{Model-only}} \\
\cmidrule(lr){7-11}
& \textbf{(m)} & & \textbf{(Y/N)} & \textbf{(m)} & \textbf{(m)} & \textbf{1s} & \textbf{2s} & \textbf{3s} & \textbf{4s} & \textbf{5s} & \textbf{(m)} & \textbf{(K)} & \textbf{(ms)} & \textbf{(ms)} \\
\midrule
1  & 20 &  8 & N & 3.24 & 6.62 & 1.15 & 2.36 & 3.67 & 5.09 & 6.62 & 3.78 & 134.5 & 3.45 & 2.09 \\
2  & 20 & 12 & N & 2.25 & 5.17 & 0.64 & 1.51 & 2.59 & 3.85 & 5.29 & 2.78 & 134.5 & 3.35 & 2.16 \\
\rowcolor{latblue}
\textbf{3}  & \textbf{20} & \textbf{16} & \textbf{N} &
\textbf{2.13} & \textbf{4.93} &
\textbf{0.59} & \textbf{1.40} & \textbf{2.42} & \textbf{3.63} & \textbf{5.01} &
\textbf{2.61} & \textbf{134.5} & \textbf{3.17} & \textbf{2.08} \\
4  & 30 &  8 & N & 2.68 & 5.92 & 0.85 & 1.88 & 3.09 & 4.49 & 6.03 & 3.27 & 134.5 & 3.69 & 2.12 \\
5  & 30 & 12 & N & 2.15 & 4.99 & 2.14 & 4.33 & 6.59 & 8.93 & 11.38 & 6.67 & 134.5 & 3.63 & 2.11 \\
6  & 30 & 16 & N & 4.58 & 5.61 & 1.74 & 3.54 & 5.42 & 7.39 & 9.46 & 5.51 & 134.5 & 3.50 & 2.13 \\
7  & 40 &  8 & N & 2.13 & 4.96 & 0.60 & 1.40 & 2.42 & 3.65 & 5.07 & 2.63 & 134.5 & 3.71 & 2.08 \\
8  & 40 & 12 & N & 2.30 & 5.24 & 0.70 & 1.59 & 2.67 & 3.94 & 5.39 & 2.86 & 134.5 & 3.71 & 2.08 \\
9  & 40 & 16 & N & 2.77 & 6.07 & 0.89 & 1.96 & 3.22 & 4.65 & 6.23 & 3.39 & 134.5 & 3.71 & 2.12 \\
\midrule
10 & 20 &  8 & Y & 2.04 & 4.85 & 0.52 & 1.29 & 2.28 & 3.49 & 4.88 & 2.49 & 145.9 & 4.57 & 3.18 \\
11 & 20 & 12 & Y & 1.89 & 4.37 & 0.56 & 1.27 & 2.16 & 3.25 & 4.54 & 2.36 & 145.9 & 4.33 & 3.14 \\
\rowcolor{tradeyellow}
\textbf{12} & \textbf{20} & \textbf{16} & \textbf{Y} &
\textbf{1.89} & \textbf{4.37} &
\textbf{0.56} & \textbf{1.26} & \textbf{2.15} & \textbf{3.24} & \textbf{4.52} &
\textbf{2.35} & \textbf{145.9} & \textbf{4.30} & \textbf{3.20} \\
13 & 30 &  8 & Y & 1.88 & 4.38 & 0.54 & 1.24 & 2.13 & 3.22 & 4.51 & 2.33 & 145.9 & 4.76 & 3.16 \\
14 & 30 & 12 & Y & 1.89 & 4.35 & 0.59 & 1.31 & 2.21 & 3.30 & 4.58 & 2.40 & 145.9 & 4.63 & 3.12 \\
\rowcolor{errgreen}
\textbf{15} & \textbf{30} & \textbf{16} & \textbf{Y} &
\textbf{1.85} & \textbf{4.25} &
\textbf{0.60} & \textbf{1.31} & \textbf{2.19} & \textbf{3.25} & \textbf{4.51} &
\textbf{2.37} & \textbf{145.9} & \textbf{4.58} & \textbf{3.21} \\
16 & 40 &  8 & Y & 1.91 & 4.42 & 0.55 & 1.26 & 2.16 & 3.26 & 4.55 & 2.36 & 145.9 & 4.81 & 3.18 \\
17 & 40 & 12 & Y & 1.89 & 4.38 & 0.56 & 1.27 & 2.16 & 3.26 & 4.54 & 2.36 & 145.9 & 4.84 & 3.16 \\
\bottomrule
\end{tabular}%
}
\end{table*}

\begin{table*}[t]
\centering
\caption{TCN-based temporal encoder variants on NGSIM. Errors are in meters; latency is in milliseconds.  Missing entries are denoted by \textbf{--}.}
\label{tab:supp_tcn_full}
\resizebox{0.8\textwidth}{!}{%
\begin{tabular}{c c c c c c c|c c|c c c c c|c c c}
\toprule
\textbf{ID} &
\textbf{$r$} &
\textbf{$K$} &
\shortstack{\textbf{Residual}\\\textbf{(Y/N)}} &
\textbf{L} &
\textbf{H} &
\shortstack{\textbf{Smooth}\\\textbf{(Y/N)}} &
\textbf{ADE} & \textbf{FDE} &
\multicolumn{5}{c|}{\textbf{RMSE (m)}} &
\shortstack{\textbf{AVG}\\\textbf{(m)}} &
\shortstack{\textbf{Params}\\\textbf{(K)}} &
\shortstack{\textbf{E2E}\\\textbf{(ms)}} \\
\cmidrule(lr){10-14}
& \textbf{(m)} & & & & & & \textbf{(m)} & \textbf{(m)} &
\textbf{1s} & \textbf{2s} & \textbf{3s} & \textbf{4s} & \textbf{5s} &
& & \\
\midrule
1  & 35 & \textbf{--} & N & 2 & 2 & N & 1.86 & 4.23 & 0.58 & 1.28 & 2.15 & 3.21 & 4.46 & 2.336 & 145.9 & \textbf{--} \\
2  & 35 & 8  & Y & 2 & 2 & N & 1.89 & 4.38 & 0.55 & 1.25 & 2.14 & 3.23 & 4.51 & 2.34  & 145.9 & 4.20 \\
3  & 35 & 12 & Y & 2 & 2 & N & 1.86 & 4.28 & 0.55 & 1.24 & 2.11 & 3.17 & 4.43 & 2.30  & 145.9 & 3.00 \\
4  & 35 & 16 & Y & 2 & 2 & N & 1.86 & 4.27 & 0.57 & 1.28 & 2.16 & 3.23 & 4.49 & 2.35  & 145.9 & 3.00 \\
\midrule
5  & 35 & 12 & Y & 8 & 4 & N & 1.84 & 4.20 & 0.56 & 1.23 & 2.08 & 3.12 & 4.34 & 2.27 & 147.9 & 6.30 \\
6  & 35 & 12 & Y & 4 & 4 & N & 1.86 & 4.29 & 0.55 & 1.25 & 2.12 & 3.18 & 4.44 & 2.31 & 146.6 & 4.50 \\
7  & 35 & 12 & Y & 4 & 2 & N & 1.86 & 4.31 & 0.56 & 1.25 & 2.13 & 3.20 & 4.47 & 2.32 & 146.6 & 4.40 \\
8  & 35 & 12 & Y & 8 & 2 & N & 1.98 & 4.55 & 0.64 & 1.36 & 2.28 & 3.43 & 4.78 & 2.50 & 147.9 & 5.00 \\
\midrule
9  & 45 & 12 & Y & 8 & 4 & Y & 1.82 & 4.26 & 0.51 & 1.21 & 2.10 & 3.18 & 4.44 & 2.29 & 147.9 & 6.30 \\
10 & 35 & 12 & Y & 8 & 4 & Y & 1.87 & 4.30 & 0.56 & 1.26 & 2.14 & 3.21 & 4.47 & 2.33 & 147.9 & 6.40 \\
11 & 35 & 12 & Y & 2 & 2 & Y & 1.87 & 4.29 & 0.57 & 1.27 & 2.15 & 3.22 & 4.47 & 2.34 & 145.9 & 3.60 \\
12 & 45 & 12 & Y & 2 & 2 & Y & 1.88 & 4.32 & 0.56 & 1.27 & 2.15 & 3.23 & 4.49 & 2.34 & 145.9 & 3.50 \\
\bottomrule
\end{tabular}%
}
\end{table*}

Table~\ref{tab:supp_ablation_full} provides the complete grid. From this sweep, we select three representative operating points for the main paper: (i) the lowest-latency configuration (EdgeVTP$_{\text{Lat}}$), (ii) the best error rates configuration (EdgeVTP$_{\text{Error}}$), and (iii) a knee-point trade-off configuration (EdgeVTP$_{\text{TF}}$) that balances sub-5ms inference with strong prediction quality. The selected configurations are summarized in Table~\ref{tab:supp_ablation_selected}. For convenience, the three operating points used in the main paper are summarized in Table~\ref{tab:supp_ablation_selected}. We note a performance degradation in IDs 5 and 6; we attribute this to increased social noise from distant neighbors in these specific configurations, suggesting a threshold where wider context becomes counter-productive for this architecture.

\subsection{Latency Breakdown: Model Inference vs.\ Pipeline Overhead}
\label{sec:latency_breakdown}

A central motivation of EdgeVTP is that, on edge hardware,
end-to-end latency is not dominated by learned-parameter
inference alone. To quantify this, Table~\ref{tab:supp_ablation_full}
decomposes the E2E latency into model-only inference and pipeline overhead, where the latter
includes graph construction (Edge Builder) and B\'ezier curve evaluation for trajectory reconstruction. Across the three selected
operating points, pipeline overhead accounts for
25-35\% of total E2E runtime. Notably, this overhead is
largely independent of whether residual separation is enabled (compare IDs 3 and 12: overhead is ${\sim}1.1$\,ms in both cases),
because it is driven by scene-dependent neighbor search and analytic curve evaluation rather than learned-parameter forward passes. When the interaction radius increases from 20\,m to 40\,m, the overhead grows further (e.g., ID\,7: 1.63\,ms, 43.9\% of E2E), confirming that graph construction cost scales with the candidate neighbor set and can become the dominant latency component in  larger-radius configurations. These findings support the architectural design choices in EdgeVTP: bounding interaction complexity via radius gating and a hard top-$K$ cap directly reduces the fastest-growing component of end-to-end latency, an effect that post-hoc model compression (e.g., pruning or quantization of learned weights) would not address.

\section{Effect of TCN Temporal Encoder}
\label{sec:supp_tcn}

We additionally evaluated a non-causal TCN temporal encoder as a drop-in replacement for the FC temporal projection to test whether temporal convolutions improve the error-latency frontier. We keep interaction graph construction and one-shot B\'ezier decoding fixed, and vary decoder capacity (layers/heads), neighbor cap $K$, residual separation, and optional smoothing. Table~\ref{tab:supp_tcn_full} reports the evaluated configurations.

\begin{table}[!t]
\centering
\caption{Jetson Nano latency (batch=1), reported in ms (lower is better).}
\label{tab:supp_nano_power_modes}
\resizebox{0.9\columnwidth}{!}{%
\begin{tabular}{lcc}
\toprule
\textbf{Model} & \textbf{10W} & \textbf{5W} \\
\midrule
STA-LSTM \cite{lin2021stalstm}                & 51.82   & 102.85 \\
VT-Former$_{LH}$ \cite{pazho2024vtformer}       & 1034.26 & 1669.51 \\
\midrule
\rowcolor[HTML]{D9D9D9}
EdgeVTP$_{\text{Lat}}$   & 27.87 & 48.46 \\
\rowcolor[HTML]{D9D9D9}
EdgeVTP$_{\text{TF}}$    & 38.27 & 66.46 \\
\rowcolor[HTML]{D9D9D9}
EdgeVTP$_{\text{Error}}$ & 37.36 & 65.52 \\
\bottomrule
\end{tabular}%
}
\end{table}

\section{Extended Jetson Power-Mode Profiling}
\label{sec:supp_jetson_profiles}

\begin{table*}[t]
\centering
\caption{Jetson Xavier NX latency (batch=1), reported in ms across power modes and CPU core configurations (lower is better).}
\label{tab:supp_xavier_power_modes}
\resizebox{0.95\textwidth}{!}{%
\begin{tabular}{lccc|ccc|ccc}
\toprule
& \multicolumn{3}{c|}{\textbf{10W}} & \multicolumn{3}{c|}{\textbf{15W}} & \multicolumn{3}{c}{\textbf{20W}} \\
\cmidrule(lr){2-4}\cmidrule(lr){5-7}\cmidrule(lr){8-10}
\textbf{Model} &
\textbf{2-core} & \textbf{4-core} & \textbf{Desktop} &
\textbf{2-core} & \textbf{4-core} & \textbf{6-core} &
\textbf{2-core} & \textbf{4-core} & \textbf{6-core} \\
\midrule

STA-LSTM \cite{lin2021stalstm}          & 29.49  & 35.37  & 23.08  & 27.79  & 30.92  & 30.52  & 23.33  & 29.82  & 30.64 \\
VT-Former$_{LH}$ \cite{pazho2024vtformer} & 416.66 & 506.78 & 328.65 & 340.34 & 429.26 & 425.49 & 334.40 & 433.84 & 400.95 \\
\midrule
\rowcolor[HTML]{D9D9D9}
EdgeVTP$_{\text{Lat}}$   & 14.06 & 16.28 & 10.85 & 11.49 & 13.68 & 13.73 & 11.85 & 14.09 & 15.66 \\
\rowcolor[HTML]{D9D9D9}
EdgeVTP$_{\text{TF}}$    & 19.10 & 22.23 & 15.34 & 15.94 & 18.76 & 19.01 & 16.21 & 19.81 & 20.22 \\
\rowcolor[HTML]{D9D9D9}
EdgeVTP$_{\text{Error}}$ & 19.07 & 21.99 & 14.89 & 15.88 & 18.81 & 18.90 & 16.23 & 19.44 & 19.69 \\
\bottomrule
\end{tabular}%
}
\end{table*}

To further characterize deployability under realistic edge constraints, we report extended batch=1 latency across additional Jetson power modes and CPU core configurations. Table~\ref{tab:supp_xavier_power_modes} profiles Jetson Xavier NX under multiple 10W/15W/20W settings with different active CPU cores, including a desktop configuration. Table~\ref{tab:supp_nano_power_modes} reports Jetson Nano under 10W and 5W modes. These measurements follow the same latency protocol used in the main paper.

Across all Xavier NX configurations, the EdgeVTP operating points remain consistently low-latency, while the VT-Former baseline is substantially slower (hundreds of milliseconds). The latency-focused EdgeVTP variant achieves the lowest runtime across power modes, and the balanced/error-focused variants incur only a modest increase. On Jetson Nano, EdgeVTP remains within tens of milliseconds under both 10W and 5W, whereas VT-Former requires seconds per inference.

\end{document}